\documentclass[journal]{IEEEtran}
\usepackage[hidelinks]{hyperref}
\usepackage{subfiles}
\usepackage{ctable}
\usepackage{amsmath}
\usepackage{bbm}
\usepackage{amssymb}
\usepackage{pifont}
\usepackage{amsthm}
\usepackage{algorithm}
\usepackage{enumitem}
\usepackage{mathtools}
\usepackage{algpseudocode}
\newtheorem{proposition}{Proposition}
\newtheorem{lemma}{Lemma}
\theoremstyle{definition}
\newtheorem{definition}{Definition}
\newtheorem{assumption}{Assumption}
\newtheorem{remark}{Remark}
\usepackage{amsmath,amsfonts}
\usepackage{array}
\usepackage{color,soul}
\usepackage[caption=false,font=normalsize,labelfont=sf,textfont=sf]{subfig}
\usepackage{textcomp}
\usepackage{url}
\usepackage{verbatim}
\usepackage{graphicx}
\def\BibTeX{{\rm B\kern-.05em{\sc i\kern-.025em b}\kern-.08em
    T\kern-.1667em\lower.7ex\hbox{E}\kern-.125emX}}
\usepackage{balance}
\usepackage{cite}

\usepackage{booktabs}
\IEEEpubid{\begin{minipage}{\textwidth}\ \\[35pt]
\textit{© 20XX IEEE.  Personal use of this material is permitted.  Permission from IEEE must be obtained for all other uses, in any current or future media, including reprinting/republishing this material for advertising or promotional purposes, creating new collective works, for resale or redistribution to servers or lists, or reuse of any copyrighted component of this work in other works.}
\end{minipage}}
\begin{document}
\title{Prognostic Framework for Robotic Manipulators Operating Under Dynamic Task Severities}

\author{Ayush Mohanty\IEEEauthorrefmark{1}, Jason Dekarske\IEEEauthorrefmark{2}, Stephen K. Robinson\IEEEauthorrefmark{2}, Sanjay Joshi\IEEEauthorrefmark{2}, Nagi Gebraeel\IEEEauthorrefmark{1}
\thanks{\IEEEauthorblockA{\IEEEauthorrefmark{1}School of Industrial and Systems Engineering,
Georgia Institute of Technology, Atlanta, GA 30332, USA}\\
\IEEEauthorblockA{\IEEEauthorrefmark{2}Department of Mechanical and Aerospace Engineering, University of California Davis, Davis, CA 95616, USA}}}

\markboth{Journal of \LaTeX\ Class Files,~Vol.~18, No.~9, September~2020}%
{How to Use the IEEEtran \LaTeX \ Templates}

\maketitle

\begin{abstract}
Robotic manipulators are critical in many applications but are known to degrade over time. This degradation is influenced by the nature of the tasks performed by the robot. Tasks with higher severity, such as handling heavy payloads, can accelerate the degradation process. One way this degradation is reflected is in the position accuracy of the robot's end-effector. In this paper, we present a prognostic modeling framework that predicts a robotic manipulator's Remaining Useful Life (RUL) while accounting for the effects of task severity. Our framework represents the robot's position accuracy as a Brownian motion process with a random drift parameter that is influenced by task severity. The dynamic nature of task severity is modeled using a continuous-time Markov chain (CTMC). To evaluate RUL, we discuss two approaches -- (1) a novel closed-form expression for Remaining Lifetime Distribution (RLD), and (2) Monte Carlo simulations, commonly used in prognostics literature. Theoretical results establish the equivalence between these RUL computation approaches. We validate our framework through experiments using two distinct physics-based simulators for planar and spatial robot fleets. Our findings show that robots in both fleets experience shorter RUL when handling a higher proportion of high-severity tasks.

\end{abstract}


\section{Introduction}\label{Intro}
\IEEEPARstart{I}{ndustrial} robots performing repetitive tasks degrade over time. Degradation can compromise the robot's ability to perform certain tasks, especially those that require high accuracy while manipulating heavy objects.  Therefore, it is important to examine how a robot's degradation depends on its operating conditions dictated here by the set of tasks that the robot performs. We examine a scenario where a fleet of autonomous robots is expected to perform different tasks that vary in their level of severity. In our experiments, severity refers to the ``weight of the payload'' manipulated by the robot while performing a task. Thus, the heavier the payload, the more severe the task and the faster the rate of degradation. 

In this paper, we develop a prognostic modeling framework to predict the remaining useful life (RUL) of a robot while accounting for the severity of the tasks assigned to the robot and how they impact the degradation of the robot. Analyzing the effect of different tasks on the robot's degradation requires constant monitoring of the robot's state of health. This may not always be practical or feasible. Additionally, data collected from a robot performing different types of tasks will likely be heterogeneous and comprised of multiple subgroups of data, each exhibiting distinct statistical characteristics.  To address this issue, we propose a data collection scheme, which utilizes a pre-specified set of calibration tasks that a robot is expected to perform periodically, i.e. every specific number of task cycles. This is analogous to implementing a periodic inspection scheme where only the data collected during those inspections is used to model a component's degradation. This scheme ensures that the data used to model the robot's degradation is homogeneous. The calibration tasks can be designed by subject matter experts to thoroughly evaluate the capabilities of the robot's end-effector without interfering with its natural degradation process. 

\begin{figure}
\includegraphics[width=\columnwidth]{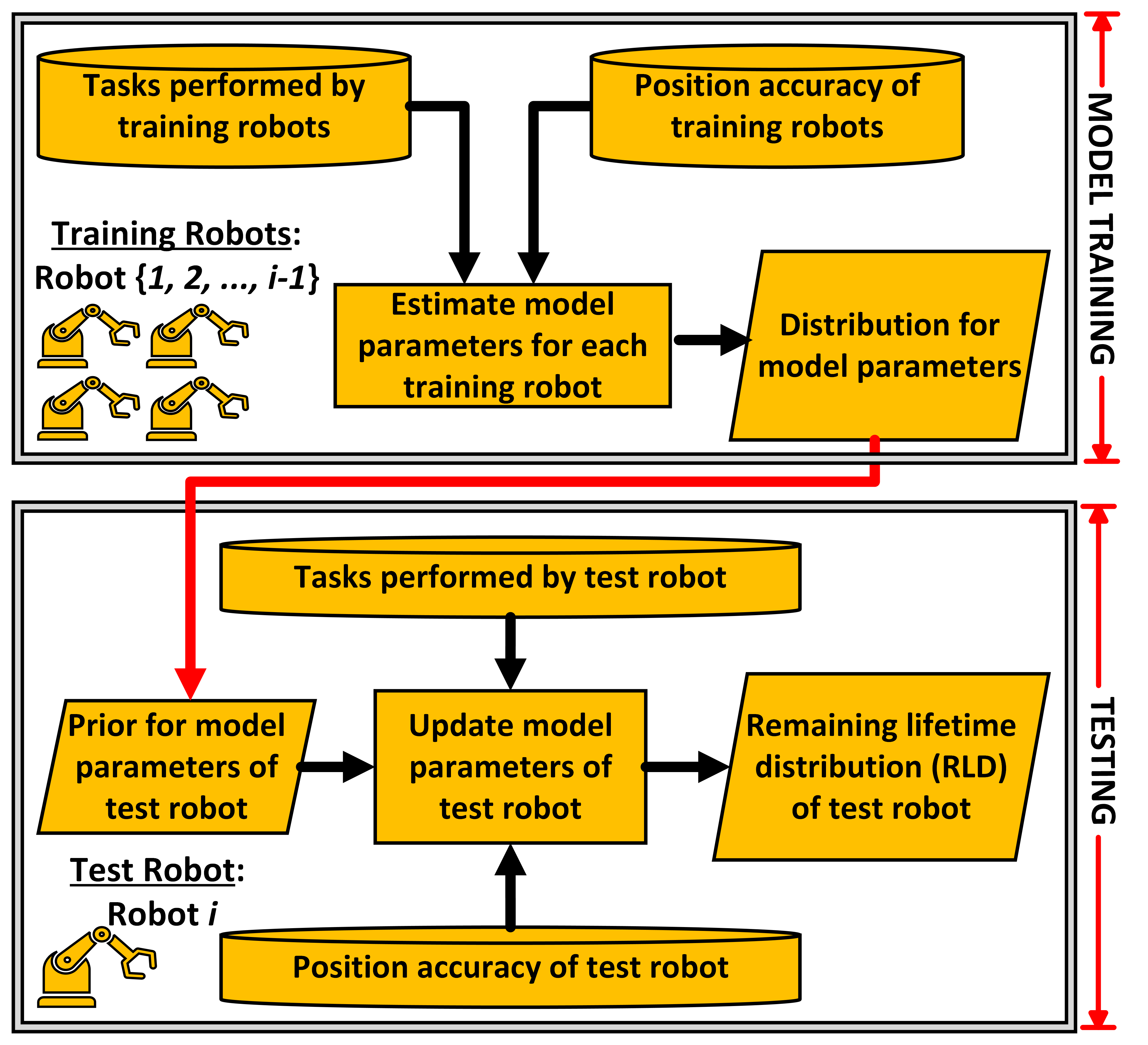}
\centering
\vspace{-0.75cm}
\caption{The proposed prognostic framework for robotic manipulators}
\label{fig:methodflowchart}
\vspace{-0.5cm}
\end{figure}

We assume that degradation occurs gradually over multiple inspection epochs. We define a robot's \textit{lifetime} as the time it takes a degradation signal to reach a pre-specified failure threshold. The threshold represents a \textit{soft failure}, i.e., a point at which the robot is incapable of successfully completing a prescribed set of tasks with the intended accuracy. We consider the \textbf{position accuracy of the end-effector as our degradation signal}. The position accuracy is defined as the deviation of the end-effector's position from a desired target, which is assumed to increase with progressive degradation. It is worth mentioning that position accuracy has been commonly used in numerous studies such as \cite{Qiao_2019_CASE}, \cite{Qiao_2021_ICRA}, \cite{Qiao_Weiss_2018_journal}, and \cite{Qiao_Weiss_2019_journal} to evaluate robot's state-of-health. 

\textbf{Main Contributions.} The key contribution of this paper centers on a prognostics framework that potentially enables the robot to decide on which tasks it can (or cannot) complete. By linking task severity to the robot’s position accuracy (degradation), this framework establishes a foundation for predicting the remaining useful life (RUL) of robots and understanding their capability under varying operational conditions. The methodology developed in this paper is data-driven, and hence, it is agnostic to the kinematics of the robot. The main technical contributions of this paper are as follows:
\begin{enumerate}
    \item We model the task severity as a Continuous-Time Markov Chain (CTMC), which captures the stochastic nature of task planning. The robot's position accuracy (degradation) is modeled using a Brownian motion process with a random drift. Furthermore, the drift is modeled to be influenced by the task severity.
    \item Using Bayesian inference we propose posterior distributions for the above models. This is intended to capture the latest degradation characteristics of the robot, and in turn, update the residual life distribution (RLD) accordingly.
    \item Statistical properties of CTMC, and Brownian motion processes are utilized to propose a novel closed-form RLD estimation approach, which is unique to the prognostics literature.  An alternative approach using Monte-Carlo (MC) simulation is also mentioned to compare the RLD estimation with the closed-form approach.
    \item We provide theoretical results proving that the closed-form RLD estimation approach converges to the MC simulations approach under mild assumptions. 
    \item Physics-based simulation experiments are conducted on two distinct robot fleets -- planar and spatial, each comprising 25 robots, to assess the robustness of the framework to \textbf{(i)} unit-to-unit variability, and \textbf{(ii)} type of robot (including differences in dynamics, controller, etc).
    \item Finally, we demonstrate the practical usefulness of our closed-form RLD estimation approach through what-if analyses, by empirically studying how variations in future task proportions influence the lifetime of the robots.
\end{enumerate}
A flowchart depicting the process of predicting the RUL of a robot is shown in Figure \ref{fig:methodflowchart}.

The rest of the article is organized as follows. Section \ref{lit_review} discusses the related literature in prognostics. A degradation modeling framework for the position accuracy is presented in Section \ref{modeling}.  Two approaches for estimating the RUL of the robots are presented in Section \ref{distribution}. Section \ref{Exper} presents the experiments performed using case studies of planar and spatial robotic manipulators, whose results are discussed in section \ref{results}. The impact of task proportions on robots' lifetime is presented in Section \ref{whatif}.

\section{Literature Review}\label{lit_review}
\subsection{Robot's Faults Diagnostics and Prognostics}
The majority of the modeling efforts that have explored the degradation of robotic systems have focused on fault diagnosis. Only very few have examined prognostic models for robots. Almost all diagnostic models have focused on specific components of the robot. For example, \cite{van2013robust}, \cite{caccavale2012discrete} focused on the diagnosis of robotic manipulators, \cite{wu2020review} provided a literature review on fault diagnosis methods that focused on robot joint servo systems, and \cite{skoundrianos2004finding} studied fault diagnosis of the wheels of a mobile robot. Unlike fault diagnosis, prognostic models aim to predict the future degradation of a robot and estimate its remaining functional lifetime. The summary of the literature reviewed in this paper, along with the identified research gap is provided in Table \ref{tab:litreview1}. 

Kinematic models that describe a robot's motion have been used to model degradation. \cite{Qiao_Weiss_2018_journal} and \cite{Qiao_Weiss_2019_journal} studied the position and orientation accuracy of a robot system's tool center position (TCP) used in manufacturing applications. The authors explored robots used in manufacturing applications. The authors developed a methodology, known as quick health assessment, to assess a robot’s TCP accuracy degradation. They developed a Fixed-Loop Measurement Plan to generate limited sample measurements to assess the robot accuracy degradation in the overall robot measurement volume. The measurement plan was based on a kinematic model to estimate a robot's position-dependent joint errors. This approach is similar in spirit to our concept of periodic inspection using calibrating tasks.

Various regression algorithms were used by \cite{Izagirre_Andonegui_Eciolaza_Zurutuza_2021} to predict the end-effector's position of a robot moving in a fixed trajectory with different payloads. However, the payload masses that were being moved by the robot were the same for each experiment. The robot's accuracy was analyzed separately for each payload. While all the aforementioned papers utilized the end-effector's accuracy as a degradation signal, none of these research efforts presented a systematic framework for computing the RUL of a robot.  

To the best of our knowledge, the RUL modeling in robots taking into account their operating conditions was formalized only by these two papers: \cite{Qibo2021} and \cite{Aivaliotis_Arkouli_Georgoulias_Makris_2021}. \cite{Qibo2021} developed a two-stage approach for detecting faults in an industrial robot and then predicting its RUL. RUL prediction considering robot-to-robot variations was evaluated using a domain-generalized long short-term memory (LSTM) model. The paper also considered dynamic working regimes by changing joint speed and torque profiles. However, it did not provide a formal relationship between the working regime and the robot's RUL. A framework by \cite{Aivaliotis_Arkouli_Georgoulias_Makris_2021} studied the degradation of an industrial manipulator in a manufacturing setting. The production plan was comprised of tasks with different combinations of payloads and trajectories. The friction coefficient of a gearbox was selected as the degradation signal of the study. The paper predicted the expected RUL of the component under study, however, the uncertainty around prediction was not well quantified. This is because the RUL prediction algorithm ran only once and assumed a deterministic degradation process. 

\begin{table}
\caption{Literature on robot's faults diagnostics and prognostics}
\centering
\setlength{\tabcolsep}{2pt}
\begin{tabular}{*5c}
\toprule
Lit. & Analysis type & Monitoring & Dynamic tasks & RUL depen. on tasks\\ 
\midrule
\cite{wu2020review} & Diagnostics & Joints' servo & \ding{55} & --\\
\cite{Qiao_Weiss_2018_journal}, \cite{Qiao_Weiss_2019_journal} & Diagnostics & End effector  & \ding{55} & --\\
\cite{Izagirre_Andonegui_Eciolaza_Zurutuza_2021} &  Diagnostics & End effector & \ding{51} & --\\
\cite{Aivaliotis_Arkouli_Georgoulias_Makris_2021} & Prognostics & Gear box & \ding{51} & \ding{55}\\
\cite{Qibo2021} & Prognostics & End effector & \ding{51} & \ding{55}\\
\textbf{Our} & Prognostics & End effector & \ding{51} & \ding{51}\\
\bottomrule
\end{tabular}
\label{tab:litreview1}
\vspace{-0.5cm}
\end{table}

\subsection{Prognostics Under Dynamic Operating Conditions}
Prognostic models leverage information about the underlying degradation process to predict the RUL of partially degraded components. Many models developed in the literature such as \cite{Deutsch2018}, \cite{Gebraeel2005}, and \cite{Zhang2022}, assume that the environmental and/or operating conditions remain constant over time. This assumption does not necessarily hold in many real-world settings. It is not unreasonable to assume that more severe operating conditions will tend to accelerate the degradation rate of a component. Therefore, the RUL predictions must account for changes in the operating conditions. Some research efforts such as \cite{Jianhui2008} and \cite{Bian2013} developed prognostic models that utilized a deterministic profile for future operating conditions. However, in the real-world, the operating conditions are often found to be stochastic. In this section, we review some of the works that focused on addressing this problem. We also identify the key methodology in each of these works and summarize them in Table \ref{tab:litreview2}. 

Prognostics models that account for dynamic operating conditions have been developed using physics-based frameworks and data-driven methodologies. Choosing which approach to use often depends on the type of application, the availability of data, and the existence of a well-studied physics model. For example, a physics-based prognostic methodology was developed by \cite{Zhao2015} to model damage propagation in a spur gear tooth. The operating conditions were assumed to evolve dynamically over time according to another physics-based model. The remaining lifetime was evaluated using an accelerated Markov Chain Monte Carlo (MCMC) algorithm. The impact of stochastic operating conditions on the degradation of various components of a subway was studied by \cite{Tamssaouet2021}. Two different operating conditions were studied -- a) the difference in the behavior of the subway's drivers and b) the differences in distances between subway stations. The RUL was computed using Monte Carlo simulations. 

In the data-driven literature, \cite{Liao2013} used a conjugate prior and an MCMC approach for non-linear degradation modeling in time-varying operating condition settings. The authors studied two case scenarios describing the operating conditions. The first assumed the operating conditions evolved in a deterministic manner that follows some mean function whereas the second case considered a stochastic environment modeled as white noise. \cite{Hong2015} used a general additive modeling framework to model degradation under dynamic operating conditions. The authors included covariates to account for trend, seasonality, and stationary in the operating conditions. The failure time in \cite{Hong2015} was computed using Monte-Carlo simulation. \cite{Kharoufeh2003} used an additive degradation model where the random environment affecting the degradation was modeled as a Continuous-Time Markov Chain (CTMC). It was further extended in \cite{Kharoufeh2005} where they applied the methodology to two distinct cases of a) observing only the environment and b) observing only the degradation state. \cite{Bian2015} used a Brownian motion process-based linear degradation model, where the drift of the Brownian motion was a function of the operating conditions. Here as well, the operating condition was modeled using a CTMC. However, \cite{Bian2015} assumed both environment and degradation state can be observed simultaneously. The remaining lifetime distribution (RLD) approximation in both \cite{Kharoufeh2005} and \cite{Bian2015} was done using two approaches -- a) a Laplace transform technique, and b) the traditional Monte-Carlo simulation. A similar problem was addressed by \cite{Jahani2021} where the operating conditions were modeled using a spline regression. They also used Monte Carlo simulation to predict RUL. 

One of the key research gaps identified in the reviewed literature was the lack of closed-form approximation for RLD when subjected to dynamic operating conditions. Addressing this research gap is the primary methodological contribution of this paper. Our methodology follows the formulation developed by \cite{Bian2015} with some key differences. First, our robot's accuracy degradation signal does not contain any shocks. This allows us to propose a closed-form expression of the RLD of the robot that still leverages the properties of the CTMC. Furthermore, our problem setting is unique in that we cannot assume that the robot is constantly being utilized. Thus, the conventional assumption used in most degradation models, i.e., monitoring at fixed time intervals will most likely generate erroneous results. In our problem setting, since a robot can sometimes be idle, we instead collect degradation (position accuracy) observations at the inspection epochs after a fixed number of operational cycles. A discussion on the data collection procedure is given in Section \ref{modeling}. 
\begin{table}
\caption{Literature on prognostics under dynamic operating conditions}
\centering
\setlength{\tabcolsep}{0.10pt}
\begin{tabular}{*4c}
\toprule
Lit. & Deg. model & Oper. Cond. model & Closed-form RLD \\
\midrule
\cite{Zhao2015} & Physics-based & Physics-based & \ding{55}\\
\cite{Tamssaouet2021} & Physics-based & Analyzed separately & \ding{55}\\
\cite{Liao2013} & MCMC & Deterministic/White-noise & \ding{55}\\
\cite{Hong2015} & Additive model & Shape-restricted splines & \ding{55}\\
\cite{Kharoufeh2003}, \cite{Kharoufeh2005} & Additive model & CTMC & \ding{55}\\
\cite{Bian2015} & BM with drift and jumps & CTMC & \ding{55}\\
\cite{Jahani2021} & BM with drift & Spline Regression & \ding{55}\\
\textbf{Our} & BM with drift & CTMC & \ding{51}\\
\bottomrule
\end{tabular}
\label{tab:litreview2}
\vspace{-0.5cm}
\end{table}

\section{Degradation Modeling Framework of Robot}\label{modeling}
\begin{figure}
\includegraphics[width=\columnwidth]{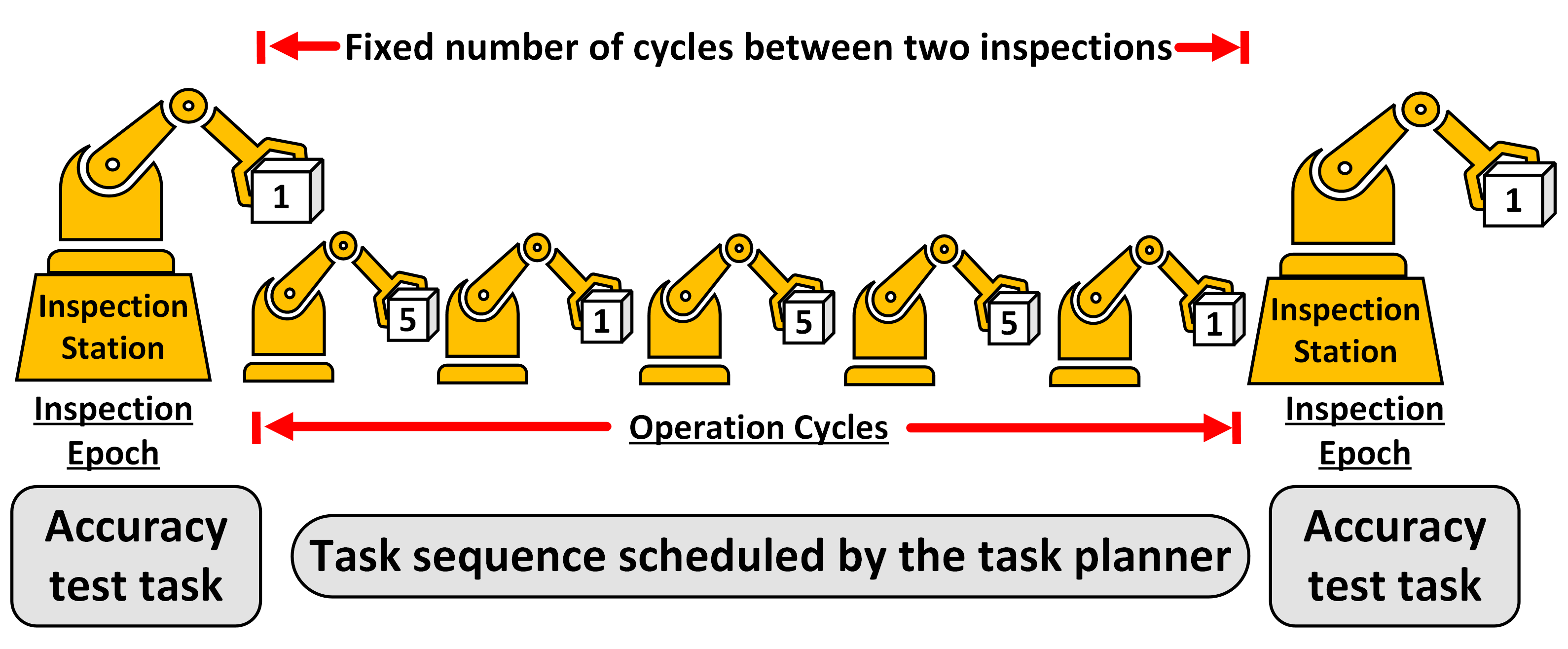}
\vspace{-0.75cm}
\caption{In the proposed framework, an accuracy test task is performed periodically to inspect the position accuracy and estimate RUL}
\label{fig:InspOps_fig}
 \vspace{-0.5cm}
\end{figure}

We model a robot's degradation by monitoring the position accuracy of its end-effector, which we assume to increase with degradation (e.g., wear) of electro-mechanical components in the robot (e.g. motors, actuators).  In practice, it may not be feasible to constantly collect position accuracy data while a robot is performing various tasks. Additionally, due to the variability of the tasks performed by the robot, any data collected while performing these tasks will be difficult to standardize.  As a result, we consider a setting that involves \textit{inspection epochs} during which a robot is periodically tested. During an inspection epoch, a robot is assigned to perform a set of calibrated \textit{accuracy test tasks} designed by a subject matter expert to evaluate its degradation. These inspection epochs are assumed to take place after a fixed number of operational cycles. Figure \ref{fig:InspOps_fig} illustrates how the proposed inspection epoch relates to the operational cycles. As evident from Figure \ref{fig:InspOps_fig}, the robot performs one task per operational cycle. {In this work, a task (or operational cycle) refers to the completion of a single functional operation assigned to the robot. Each task begins when the robot initiates motion to perform the operation and ends once the motion segment is completed. Task severity is determined by attributes of the operation, such as payload weight.} The robot must be restricted to perform a consistent set of accuracy test tasks during its inspection epoch. This is done to ensure homogeneity of the data that is collected during inspection. Furthermore, we also assume that inspection epochs are not too frequent, thus, they do not impact the robot's degradation. For example, in our experiments discussed in Section \ref{Exper}, we use operation cycles to inspection epoch ratio of 50:1, i.e., an inspection epoch is performed every 50 operational cycles. 
\begin{remark}
    A large ratio of operational cycles to inspection epochs ensures negligible degradation during inspection epochs. The use of ratio ``50:1'' is an example, and practitioners can use larger ratios. We utilized 50:1 for a reasonable computational workload in the simulations. 
\end{remark}

     \subsection{Key Modeling Assumptions} \label{assumptions}
    We now mention two key assumptions used to model the proposed prognostic framework. The next sections elaborate on the models.

\begin{assumption}\label{assumption_TaskSeverity}
    The sequence of tasks with dynamic severity follows a \textit{\textbf{Continuous Time Markov Chain (CTMC)}}. The elements of the \textit{generator matrix} (Q matrix) of this CTMC are random variables.
\end{assumption}

\begin{assumption}\label{assumption_Degradation}
    The degradation of the robot follows a \textit{\textbf{Brownian motion process}} with random drift. In addition, this drift depends on the severity of the task modeled by the CTMC (mentioned in Assumption \ref{assumption_TaskSeverity}).     
\end{assumption}

Section \ref{TS} discusses the CTMC that models the dynamic task severity. The details of the Brownian motion-based degradation model are provided in Section \ref{RobotAccuracy}.

    \subsection{Task Severity Model} \label{TS}
    We consider a setting where a \textit{task planner} generates a sequence of tasks for the robot to complete during its operational cycles (Figure \ref{fig:InspOps_fig}). The process of generating tasks with varying severity levels is modeled as a continuous-time Markov chain (CTMC). Let $ \{\psi(c): c > 0\} $ represent this CTMC, where the states correspond to distinct task severity levels. The state space of the CTMC is denoted as $ \mathcal{S} = \{1, 2, \dots\} $, and its infinitesimal generator matrix is $ Q = [q_{ij}] $, where $ q_{ij} $ represents the transition rate from severity level $ i $ to $ j $ ($ i, j \in \mathcal{S}, i \neq j $) and $ q_{ii} = -\sum_{j \neq i} q_{ij} $.

The transition rates $ q_{ij} $ are treated as random variables, referred to as the \textit{severity model parameters}. In this work, we assume $ q_{ij}, \, i \neq j $, follow a gamma prior distribution with shape and scale parameters $ k^{\text{prior}}_{ij} $ and $ \theta^{\text{prior}}_{ij} $, respectively. This assumption is consistent with prior work in degradation modeling, such as \cite{Bian2015}. In Section \ref{BayesUpdating_severity}, we demonstrate how the $ q_{ij} $ parameters can be updated using Bayesian methods, incorporating observations of tasks and their severities.

In our experiments outlined in Section \ref{TaskPlanner}, we characterize task severity based on the \textit{weights of the payloads} handled by the robotic manipulator.

   \subsection{Robot Position Accuracy Model} \label{RobotAccuracy}
    The degradation in the robotic manipulator is manifested in the position accuracy observed at the robot's end-effector, which we term as \textit{robot's position accuracy} in Definition \ref{accuracy_def}. This is also consistent with the terminology ``pose accuracy" used as a performance criterion for industrial manipulators in the ISO 9283 document \cite{ISO9283}. 
\begin{definition}\label{accuracy_def}
    A robot's position accuracy is defined as the Euclidean distance between the end-effector's actual position and the commanded position after the completion of a cycle.
\end{definition}
The Brownian motion-based degradation outlined in Assumption \ref{assumption_Degradation} is used to model the robot's position accuracy. Furthermore, Assumption \ref{assumption_Degradation} mentions that the drift of the Brownian motion is a random variable. In the context of prognostics, this drift is interpreted as the \textit{degradation rate}. This framework associates the degradation rate directly with the severities of the tasks performed by the robot, as noted in Assumption \ref{assumption_TaskSeverity}. 

The robot's position accuracy is monitored at inspection epochs using a nominal payload as discussed in the first paragraph of Section \ref{modeling}. Let $ A(c_k) $ represent the \textit{observed} position accuracy at the current epoch $ c_k $, and $ A(c_{k+i}) $ the \textit{predicted} accuracy at a future epoch $ c_{k+i} $. The, $ A(c_{k+i})$ is modeled as:
\begin{equation}\label{accuracy_linear}
    A(c_{k + i}) = A(c_k) + \int_{c_k}^{c_{k + i}} r(\psi(\nu)) d\nu + \gamma W(c_{k + i} - c_k),
\end{equation}
where $ r(\psi(.)) $ represents the degradation rate as a function of task severity $ \psi(\nu) $, and $ \gamma W(c_{k+i} - c_k) $ is a Brownian motion term with diffusion parameter $ \gamma $. The error term accounts for autocorrelated noise commonly observed in sensor data, following $ \gamma W(c) \sim \mathcal{N}(0, \gamma^2 c) $, where $ \gamma > 0 $ and $ c > 0 $.

The degradation rate $ r(\psi(\nu)) $ is assumed to vary linearly with task severity $ \psi(\nu) $, modeled as:
\begin{equation}\label{rate_accuracy}
    r(\psi(\nu)) = \alpha \psi(\nu) + \beta,
\end{equation}
where $ \alpha $ and $ \beta $ are random variables, referred to as the \textit{accuracy model parameters}. Specifically, $ \alpha \sim \mathcal{N}(\mu_1, \sigma_1^2) $ and $ \beta \sim \mathcal{N}(\mu_2, \sigma_2^2) $, representing independent normal prior distributions with parameters $ (\mu_1, \sigma_1^2) $ and $ (\mu_2, \sigma_2^2) $, respectively. Treating $ \alpha $ and $ \beta $ as random variables enhance robustness in scenarios involving unit-to-unit variability, such as managing a fleet of robots. In Section \ref{BayesUpdating_accuracy}, we detail a Bayesian approach for updating the distributions of these parameters based on observed data.

    \subsection{Bayesian Updating of Model Parameters} \label{BayesUpdating}
    \subsubsection{Update of Severity Model Parameters}\label{BayesUpdating_severity}
Let the task severities observed before the inspection epoch $c_k$ be denoted by the set ${\mathcal{P}}_{c_k}$. Given ${\mathcal{P}}_{c_k}$, we use Bayesian updating to obtain the posterior distribution of $q$ as given by Equation \ref{update_severity} where $\nu_Q(q |\mathcal{P}_{c_k})$, $\pi_Q(q)$ and $\mathit{f}(\mathcal{P}_{c_k}|q)$ are the posterior distribution, prior distribution and likelihood respectively. 
\begin{equation}\label{update_severity}
\nu_Q(q |\mathcal{P}_{c_k}) = \pi_Q(q) \times \mathit{f}(\mathcal{P}_{c_k}|q)
\end{equation}

We know from Section \ref{TS} that the severity model parameters $q_{ij}$'s follow a \textit{gamma prior distribution}. Let $n_{ij}(c_k)$ and $h_i(c_k)$ denote the observed number of transitions from states $i$ to $j$, and the observed holding time in state $i$ for the CTMC during the interval $[0, c_k]$. Utilizing Theorem 1 of \cite{Bian2015} we can say that $q_{ij}$ has a \textit{gamma posterior distribution} with shape parameters $k^{pos}_{ij}$ and scale parameters $\theta^{pos}_{ij}$, given by Equations \ref{shape_gamma} and \ref{scale_gamma} respectively. 
\begin{gather}
    k^{pos}_{ij} = k^{prior}_{ij} + n_{ij}(c_k) \label{shape_gamma} \\
    \theta^{pos}_{ij} = [{(\theta^{prior}_{ij})}^{-1} + h_i(c_k)]^{-1} \label{scale_gamma}
\end{gather}

\subsubsection{Update of Accuracy Model Parameters}\label{BayesUpdating_accuracy}
Recall that we mentioned the accuracy model parameters $\alpha$ and $\beta$ to follow two independent normal prior distributions. Let ${\mathcal{A}}_{c_k}$ denote the set of accuracy observed up to inspection epoch $c_k$. Given, ${\mathcal{A}}_{c_k}$ and ${\mathcal{P}}_{c_k}$, the accuracy model parameters are updated using Bayes' formula described by Equation \ref{bayes_acc_2} where, the terms $\nu_A(\alpha, \beta | \mathcal{P}_{c_k})$, $\pi_A(\alpha), \pi_A(\beta)$ and $\mathit{f}(\mathcal{A}_{c_k} |  \mathcal{P}_{c_k}, \alpha, \beta)$ represents the posterior distribution, prior distribution, and likelihood respectively.
\begin{equation}\label{bayes_acc_2}
    \nu_A(\alpha, \beta |\mathcal{A}_{c_k}, \mathcal{P}_{c_k}) \propto \pi_A(\alpha)\pi_A(\beta)\mathit{f}(\mathcal{A}_{c_k} | \mathcal{P}_{c_k}, \alpha, \beta)
\end{equation}

Re-parametrization allows us to derive a closed-form expression for the posterior mean and variance of the accuracy model parameters as follows. We first define $A_i = A(c_i) - A(c_{i-1})$. Without loss of generality, it is assumed that $A_1 = A(c_1)$. Similarly, let $\psi_i = \int_{c_{i - 1}}^{c_i}\psi(c) dt$ for task severities observed between two consecutive inspection epochs $c_{i-1}$ and $c_{i}$. Again we, assume that $\psi_1 = \psi(c_1)$. Using Equation \ref{accuracy_linear} and \ref{rate_accuracy} one can infer that $A_i \sim \mathcal{N}([\alpha \psi_i + \beta (c_i - c_{i-1})], \gamma^2(c_i - c_{i-1})) $. Therefore, the likelihood function of this normally distributed accuracy data is expressed as Equation \ref{acc_likelihood}.  
\begin{equation}\label{acc_likelihood}
    \mathit{f}(\mathcal{A}_{c_k} |\mathcal{P}_{c_k}, \alpha, \beta) = \prod_{i = 1}^k \phi\bigg(A(c_i) - A(c_{i-1})\bigg) = \prod_{i = 1}^k \phi(A_i)
\end{equation}
Now, given $\mathcal{A}_{c_k}$ and $\mathcal{P}_{c_k}$, we utilize Proposition \ref{prop1} to compute the posterior distribution of accuracy model parameters.
\begin{proposition}\label{prop1}
    At any inspection epoch $c_k$, the posterior distribution of $\alpha$ and $\beta$ follows a bivariate normal with mean ($\mu_{\alpha}, \mu_{\beta}$), variance ($\sigma_{\alpha}^2, \sigma_{\beta}^2$) and correlation coefficient $\rho$ with the following values: 
\end{proposition}
\small
\begin{equation}
     \mu_{\alpha} = \frac{\splitfrac{\Big(\mu_1\gamma^2 + \sigma_1^2\underset {i = 1} {\overset {k}{\sum}}\frac{(A_i)\cdot(\psi_i)}{(t_i - t_{i-1})}\Big)\Big(\gamma^2 + \sigma_2^2 t_k\Big)}{ - \Big(\sigma_1^2 \underset {i = 1} {\overset {k}{\sum}} \psi_i \Big)\Big(\mu_2\gamma^2 + \sigma_2^2\underset {i = 1} {\overset {k}{\sum}}A_i\Big)}}{\Big(\gamma^2 + \sigma_1^2 \underset {i = 1} {\overset {k}{\sum}}\frac{(\psi_i)^2}{(c_i - c_{i-1})}\Big)\Big(\gamma^2 + \sigma_2^2c_k\Big) - \sigma_1^2\sigma_2^2\Big(\underset {i = 1} {\overset {k}{\sum}}\psi_i\Big)^2}\\
\end{equation}

\begin{equation}
     \mu_{\beta} = \frac{\splitfrac{\Big(\mu_2\gamma^2 + \sigma_2^2\underset {i = 1} {\overset {k}{\sum}}A_i\Big)\Big(\gamma^2 + \sigma_1^2 \underset {i = 1} {\overset {k}{\sum}}\frac{(\psi_i)^2}{(c_i - c_{i-1})}\Big)} {-  \Big(\mu_1\gamma^2 + \sigma_1^2\underset {i = 1} {\overset {k}{\sum}}\frac{(A_i)\cdot(\psi_i)}{c_i - c_{i-1}}\Big)\Big(\sigma_2^2 \underset {i = 1} {\overset {k}{\sum}} \psi_i \Big)}}{\Big(\gamma^2 + \sigma_1^2 \underset {i = 1} {\overset {k}{\sum}}\frac{(\psi_i)^2}{(c_i - c_{i-1})}\Big)\Big(\gamma^2 + \sigma_2^2c_k\Big) - \sigma_1^2\sigma_2^2\Big(\underset {i = 1} {\overset {k}{\sum}}\psi_i\Big)^2}\\
\end{equation}

\begin{equation}
    \sigma_{\alpha}^2 = \frac{\Big(\gamma^2 + \sigma_2^2c_k\Big) \Big(\sigma_1^2\gamma^2\Big)}{\Big(\gamma^2 + \sigma_1^2 \underset {i = 1} {\overset {k}{\sum}}\frac{(\psi_i)^2}{(c_i - c_{i-1})}\Big)\Big(\gamma^2 + \sigma_2^2c_k\Big) - \sigma_1^2\sigma_2^2\Big(\underset {i = 1} {\overset {k}{\sum}}\psi_i\Big)^2}\\
\end{equation}

\begin{equation}
    \sigma_{\beta}^2 = \frac{\Big(\gamma^2 + \sigma_1^2 \underset {i = 1} {\overset {k}{\sum}}\frac{(\psi_i)^2}{(c_i - c_{i-1})}\Big) \Big(\sigma_2^2\gamma^2\Big)}{\Big(\gamma^2 + \sigma_1^2 \underset {i = 1} {\overset {k}{\sum}}\frac{(\psi_i)^2}{(c_i - c_{i-1})}\Big)\Big(\gamma^2 + \sigma_2^2c_k\Big) - \sigma_1^2\sigma_2^2\Big(\underset {i = 1} {\overset {k}{\sum}}\psi_i\Big)^2}\\
\end{equation}

\begin{equation}
    \rho = \frac{-\sigma_1\sigma_2\Big(\underset {i = 1} {\overset {k}{\sum}}\psi_i\Big)}{\sqrt{\Big(\gamma^2 + \sigma_1^2 \underset {i = 1} {\overset {k}{\sum}}\frac{(\psi_i)^2}{(c_i - c_{i-1})}\Big)\Big(\gamma^2 + \sigma_2^2c_k\Big)}}\\
    \medskip
\end{equation}
\normalsize
\begin{proof}
    Please refer to Appendix \ref{Proof_Prop1} for proof.
\end{proof}

\section{Remaining Lifetime Distribution of Robot}\label{distribution}
As discussed in Section \ref{Intro}, the \textit{lifetime} of a robot is defined as the first time when the position accuracy crosses a fixed \textit{soft failure} threshold, $D$. This threshold is often pre-specified by subject matter experts (see Remark \ref{threshold_remark}). Let  $R_k = \inf\{z \hspace{0.1cm} | \hspace{0.1cm} A(c_k + z) \geq D\}$ denote the RUL of the robot at inspection epoch $c_k$. Given the set of observed severity $\mathcal{P}_{c_k} $ and the set of observed accuracy $\mathcal{A}_{c_k}$, the RLD at inspection epoch $c_k$ can be expressed as follows:
\begin{equation}\label{def_RLD}
    P(R_k \leq C - c_k | {\mathcal{A}_{c_k}}, {\mathcal{P}_{c_k}}) = P(\sup_{c_k \leq u \leq C} A(u) \geq D | {\mathcal{A}_{c_k}}, {\mathcal{P}_{c_k}})
\end{equation}
In the following subsections, we present two approaches to estimate the RLD given in Equation \ref{def_RLD}. While Approach 1 utilizes the stationary distribution of the CTMC model to obtain an effective degradation rate for the robot, Approach 2 relies on numerical simulations.

    \subsection{Approach 1: Effective Degradation Rate} \label{Algo1}
Given the parameters of a CTMC, we compute its stationary distribution (represented by $\pi$) using Equation \ref{stat_dist}. 
\begin{equation} \label{stat_dist}
    \pi Q = 0 \hspace{0.3cm} \text{and} \hspace{0.3cm} \sum_{i \in \mathcal{S}}\pi_i = 1 
\end{equation}

It is a known fact that the first hitting time of a Brownian motion with deterministic drift follows an Inverse Gaussian distribution. However, in our case, the drift is non-deterministic due to the stochastic rate of degradation as evident in Equations \ref{accuracy_linear} and \ref{rate_accuracy}. To account for this random rate of degradation (or drift of Brownian motion), \cite{Elwany2009} used a conservative mean approach to estimate the first passage using an Inverse Gaussian distribution. 

Using the same technique, an expression for the RUL of the robot i.e., $T$ can be given by Equation \ref{T_MIG}. Furthermore, Equation \ref{T_MIG} leverages the stationary distribution of the task severity CTMC  to provide an ``\textit{effective degradation rate}" given by Equation \ref{phi}. 
\begin{equation}\label{T_MIG}
    T \sim \mathcal{IG}\bigg(\frac{D - A(c_k)}{\mu_{\phi}}, \frac{(D - A(c_k))^2}{\gamma^2}\bigg)
\end{equation}
where, 
\begin{equation}\label{phi}
    \mu_{\phi} = \sum_{i \in \mathcal{S}} \pi_i \cdot \big(\mu_{\alpha} \times \{\psi(.) = i\} + \mu_{\beta}\big)
\end{equation}
and, $\pi$ is the stationary distribution of the CTMC that models task severity. The effective degradation rate approach is mentioned in Algorithm \ref{alg:algo1}. 
\begin{remark}\label{remark_stationaryDistribution}
        In a real-world implementation, it is not necessary to compute the $Q$ matrix or solve Equation \ref{stat_dist} to determine $\pi$. For a long prediction horizon, \textbf{\textit{the stationary distribution $\pi$ represents the proportions of different tasks scheduled by the task planner}}. This is further supported by Proposition \ref{prop2_CTMC} mentioned in Section \ref{theoretical}. Thus, practitioners can empirically observe the (historical) task sequence and estimate the stationary distribution of the CTMC. 
\end{remark}
\vspace{-0.25cm}
\begin{algorithm}
\caption{Effective degradation rate approach for RLD} \label{alg:algo1}
\textbf{Input:} $D, c_k, A(c_k), \mu_{\alpha}, \mu_{\beta}, q_{ij} \hspace{0.3cm} \forall i, j \in \mathcal{S}$\\
\textbf{Output:} Lifetime distribution $P$
\begin{algorithmic}
\State Solve the system of Equations $\sum_{j \in \mathcal{S}}\pi_i q_{ij} = 0$ and $\sum_{i \in \mathcal{S}} \pi_i = 1$ to compute $\pi_i \hspace{0.3cm} \forall i\in\mathcal{S}$ \Comment{Equation \ref{stat_dist}}\\
\State $\mu_{\phi} \gets \sum_{i \in S} \pi_i (\mu_{\alpha} \times i + \mu_{\beta})$ \hspace{0.3cm} \Comment{Equation \ref{phi}}\\
\State $\eta \gets \frac{D - A(c_k)}{\mu_{\phi}}$ \Comment{Mean of Inverse Gaussian}\\
\State $\zeta \gets \frac{(D - A(c_k))^2}{\gamma^2}$ \Comment{Shape parameter of Inverse Gaussian}\\
\State $P \gets \Phi\Large(\sqrt{\frac{\zeta}{c}} \large(\frac{c}{\eta} - 1\large)\Large) + exp\Large(\frac{2\zeta}{\eta}\Large) \Phi\Large(-\frac{\zeta}{c}\large(\frac{c}{\eta} + 1\large)\Large) \hspace{0.3cm} \forall i \in \mathcal{S}$ \Comment{This is the cdf of $\mathcal{IG}(\eta, \zeta)$}
\end{algorithmic}
\end{algorithm}
\vspace{-0.45cm}
    
    \subsection{Approach 2: Monte-Carlo Simulation}\label{Algo2}
    In this subsection, we discuss a numerical simulation-based method to compute the RLD of the robot. Let $C$ be the inspection epoch at which the robot's accuracy is observed to cross the threshold $D$. Then, the expression for RLD  of the robot is adapted from Theorem 1 of \cite{Wang1997} to formulate Equation \ref{BCP_piecewiselinear} as follows:
\begin{equation}\label{BCP_piecewiselinear}
    \mathbb{P}(R_{k} \leq C - c_k | \mathcal{A}_{c_k},  \mathcal{P}_{c_k}, \alpha, \beta, q) = 1 - \mathbb{E}[g]
\end{equation}
where, 
\begin{equation}\label{g}
\begin{split}
    g &= \prod_{C \geq c_{j} > c_k} \mathbbm{1} \bigg\{W(c_j) \leq \frac{ f_{k}(c_j)}{\gamma}\bigg\} \Bigg(1 - \\ & exp\bigg[-\frac{2\big(\frac{f_{k}(c_j)}{\gamma} - W(c_j)\big)\big(\frac{f_{k}(c_{j -1})}{\gamma} - W(c_{j-1})\big)}{c_j - c_{j-1}}\bigg]\Bigg)
\end{split}
\end{equation}
and, 
\begin{equation}\label{piecewiselinear}
    f_{k}(c) = D - \bigg( A(c_k) + \int_{c_k}^{c} (\alpha (\psi(\nu)) + \beta) d\nu\bigg)\hspace{0.37cm} \forall c > c_k
\end{equation}
Unlike our model, in the original theorem of \cite{Wang1997}, the coefficients of $ f_{k}(c)$ are deterministic constants. Therefore, to adapt Equation \ref{BCP_piecewiselinear} for random variables $\alpha$, $\beta$ and $q_{ij}$'s, we utilize Monte-Carlo simulation. 

We first generate $M$ sample paths for accuracy model parameters i.e., $\alpha$, $\beta$, and severity model parameters i.e., $q_{ij}$'s. Additionally, $\mathbb{E}[g]$ is also computed numerically for each sample path. For $M$ sample paths, the RLD using this approach is then given by Equation \ref{sample_av}. The Monte-Carlo simulation approach is given in Algorithm \ref{alg:algo2}. 
\begin{equation} \label{sample_av}
\begin{split}
    &\mathbb{P}(R_{k} \leq C - c_k | \mathcal{A}_{c_k},  \mathcal{P}_{c_k}) \\ &= \frac{1}{M} \sum_{m = 1}^M\mathbb{P}_m(R_{k} \leq C - c_k | \mathcal{A}_{c_k},  \mathcal{P}_{c_k}, \alpha^{(m)}, \beta^{(m)}, q^{(m)}) 
\end{split}
\end{equation}
\vspace{-0.5cm}
\begin{algorithm}
\caption{Monte-Carlo simulation approach for RLD} \label{alg:algo2}
\textbf{Input:} $M, N, c_k, D, f_k, (\mu_{\alpha}, \sigma_{\alpha}^2), (\mu_{\beta}, \sigma_{\beta}^2), q_{ij} \hspace{0.2cm} \forall i, j \in \mathcal{S}$\\
\textbf{Output:}  Lifetime distribution $P$
\begin{algorithmic}
\State $Sum_P \gets 0$ \Comment{This is a 0 vector}
\While{$m \leq M$}
\State Generate $\Psi = \{\psi(c): c\geq c_k\}$ using CTMC
\State Sample $\alpha \sim \mathcal{N}(\mu_{\alpha}, \sigma_{\alpha}^2)$ and $\beta \sim \mathcal{N}(\mu_{\beta}, \sigma_{\beta}^2)$
\State $g \gets 0$ \Comment{This is a 0 vector}
\While{$n \leq N$}
    \State Compute $g^{(n)}$ using Equation \ref{g} 
    \State $g \gets g + g^{(n)}$
\EndWhile
\State $E_g \gets \frac{g}{N}$ \Comment{Numerical computation of $\mathbb{E}[g]$}
\State $P_m \gets 1 - E_g$
\State $Sum_P \gets Sum_P + P_m$
\EndWhile
\State $P \gets \frac{Sum_P}{M}$
\end{algorithmic}
\end{algorithm}
\vspace{-0.75cm}
    
        \subsection{Implementation Procedure}\label{implementation}
    In this subsection, we provide the detailed step-by-step implementation procedure of the proposed prognostic framework. At any inspection epoch $c_k$,
\begin{itemize}
    \item \textbf{Step 1:} We observe the historical task severity sequence $\mathcal{P}_{c_k}$, and make the robot perform the \textit{calibrated accuracy test task} to record the position accuracy $\mathcal{A}_{c_k}$
    \item \textbf{Step 2:} Using $\mathcal{P}_{c_k}$, and $\mathcal{A}_{c_k}$ we update the parameters of the task severity model, and position accuracy models as follows: 
    \begin{itemize}
        \item The updated severity model parameters $k_{ij}, \theta_{ij}$ are obtained using equations \ref{shape_gamma}, and \ref{scale_gamma} 
        \begin{itemize}
            \item Updating $k_{ij}, \theta_{ij}$  involves obtaining number of transitions from $i$ to $j$ i.e., $n_{ij}(c_k)$, and holding time at state $i$ i.e., $h_i(c_k)$. They are obtained from the observation $\mathcal{P}_{c_k}$. 
        \end{itemize}
        \item  Proposition \ref{prop1} gives the posterior distribution of the position accuracy model parameters $\alpha$, and $\beta$
        \begin{itemize}
            \item Posterior distribution of $\alpha$, and $\beta$ involves obtaining $\sum_1^k \psi_i$ and $\sum_1^k A_i$ which can be computed from observed $\mathcal{P}_{c_k}$ and $\mathcal{A}_{c_k}$ respectively.   
            \end{itemize}
    \end{itemize}
    \item \textbf{Step 3:} We utilize the parameters of the posterior distribution of the severity model and the position accuracy model to obtain RLD using either Approach 1 or 2
    \begin{itemize}
        \item \textbf{If using Approach 1:}
        \begin{itemize}
            \item Compute the stationary distribution of the CTMC (task severity model) given by $\pi$
            \item Obtain the mean of effective degradation rate i.e., $\mu_{\phi}$ using eq \ref{phi}
            \item Compute the parameters of the inverse Gaussian distribution using eq \ref{T_MIG}
            \item The cdf of the inverse Gaussian distribution is the RLD of the robot
        \end{itemize}
        \item \textbf{If using Approach 2:}
        \begin{itemize}
                \item First compute $\mathbb{E}[g]$ separately using the following steps: 
            \begin{itemize}
                \item Generate a sample path for the CTMC $\psi$
                \item Sample the accuracy model parameters $\alpha$ and $\beta$ from their posterior distribution. 
                \item Compute $f_k(c)$ for all future cycles $c > c_k$ using eq \ref{piecewiselinear}
                \item Use $f_k(c)$s to obtain $g$ using eq \ref{g}
                \item Repeat the above steps for large sample paths ($\geq 10,000$). The sample average of $g$ provides an estimate of $\mathbb{E}[g]$
            \end{itemize}
            \item Generate a sample path for the CTMC $\psi$ and sample the accuracy model parameters $\alpha$ and $\beta$ from the posterior distribution 
            \item Obtain the RLD for that sample-path using eq \ref{BCP_piecewiselinear}
            \item Repeat the above steps for a large number of sample paths ($\geq 10,000$) and generate RLDs for each sample path. The sample average of these RLDs estimates the RLD of the robot
        \end{itemize}
    \end{itemize}
\end{itemize}

Readers are encouraged to go through the Algorithms \ref{alg:algo1} and \ref{alg:algo2} for the pseudocodes of both Approach 1 and 2 respectively. 

    \subsection{Theoretical Analysis}\label{theoretical}
    In this subsection, we show a theoretical result (Lemma 1) highlighting the relationship of the lifetime distribution computed using Approaches 1 and 2. We first start with an assumption on the task severity model (CTMC) and use a result from \cite{serfozo2009}. 

\begin{assumption}\label{assumption_ergodic}
    The CTMC is \textit{ergodic} with a unique stationary distribution. 
\end{assumption}
{Ergodicity in the context of a CTMC implies that the chain is \textit{irreducible} (every state can be reached from any other state) and \textit{positive recurrent} (the expected return time to each state is finite). This guarantees that the state probabilities converge to a unique stationary distribution. In our setting, Assumption} \ref{assumption_ergodic} {is reasonable because industrial task planners, by design, schedule a recurring mix of tasks over extended horizons, ensuring that all task severity states are revisited over time. This leads to stable long-run task proportions, consistent with the stationary distribution of the CTMC.}
\begin{proposition}[adapted from \cite{serfozo2009}] \label{prop2_CTMC}
    For an ergodic CTMC $\psi$ with stationary distribution $\pi$, we have  
\begin{equation}
    \lim_{c \to \infty} \frac{1}{c} \cdot \int_{0}^{c} \psi(\nu) d\nu = \mathbb{E}_{\pi}[\psi] = \sum_{i \in S} \pi_i \times \{\psi(.) = i\} 
\end{equation}
\end{proposition}
\begin{proof}
    Please refer to the Appendix \ref{Proof_Prop2} for proof.
\end{proof}

Proposition 2 states that the long run time average of an ergodic Markov chain is the same as its expectation w.r.t. its stationary distribution. Readers are encouraged to read the Appendix for the proof of Proposition \ref{prop2_CTMC}. Next, we define the two estimates of robot RUL done using Approaches 1 and 2 respectively. 
\begin{definition}\label{def1}
    We let $T_1$ denote the RUL computed using the effective degradation rate (approach 1). Then, 
\begin{equation}
    T_1 \sim \mathcal{IG}\bigg(\frac{D - A(c_k)}{\sum_{i \in S} \pi_i \cdot \big(\mu_{\alpha} \cdot \{\psi(.) = i\} + \mu_{\beta}\big)}, \frac{(D - A(c_k))^2}{\gamma^2}\bigg)
\end{equation}
\end{definition}
\begin{definition}\label{def2}
    We let $T_2$ denote the RUL computed using the Monte Carlo simulations (Approach 2). It must be noted that $T_2$ is a function of $M$ i.e., the number of sample paths, and $(C - c_k)$ i.e., the length of the prediction horizon. Then,  
\begin{equation}
\begin{split}
        T_2\big(M, &(C - c_k)\big) \sim \\ &\frac{1}{M} \sum_{m = 1}^M\mathbb{P}_m(R_{k} \leq C - c_k | \mathcal{A}_{c_k},  \mathcal{P}_{c_k}, \alpha^{(m)}, \beta^{(m)}, q^{(m)}) 
\end{split}
\end{equation}
\end{definition}
\begin{lemma}
    At any inspection epoch $c_k$, 
\begin{equation}\label{lemma_Equation}
        \mathbb{P}\Bigg(\lim_{(C - c_k) \to \infty}\bigg[\lim_{M \to \infty} T_2\big(M, (C - c_k)\big)\bigg] \geq \mathbb{E}[T_1]\Bigg) = 1
    \end{equation} 
\end{lemma}
where, $T_1$ and $T_2$ are as defined by definitions \ref{def1} and \ref{def2} respectively. 
\begin{proof}
    First, by the strong law of large numbers, we can say that, 
\begin{equation}\label{proof_lemma1_exp1}
    \mathbb{P}\Big(\lim_{M \to \infty} T_2(M, (C - c_k)) = \mathbb{E}[T_2]\Big) = 1
\end{equation}
We then utilize Equation \ref{sample_av} to compute $\mathbb{E}[T_2]$ as follows, 
\begin{equation}
    \mathbb{E}[T_2] = \mathbb{E}[\mathbb{P}_m(R_{k} \leq C - c_k | \mathcal{A}_{c_k},  \mathcal{P}_{c_k}, \alpha^{(m)}, \beta^{(m)}, q^{(m)})] 
\end{equation}
    
Now $\mathbb{P}_m(R_{k} \leq C - c_k | \mathcal{A}_{c_k},  \mathcal{P}_{c_k}, \alpha^{(m)}, \beta^{(m)}, q^{(m)})$ is the lifetime distribution of the following Brownian motion:

\begin{equation}
\begin{split}
D = A(c_k) +  \Bigg(\frac{\int_{c_k}^C \alpha^{(m)} \psi(\nu) + \beta^{(m)} d\nu}{(C - c_k)}\Bigg) & \times (C - c_k) \\ &+ \gamma W(C - c_k)
\end{split}
\end{equation}

Utilizing same procedure as \cite{Elwany2009}, we can say that $T_2$ is an Inverse gaussian random variable such that, 
\begin{equation}\label{proof_lemma1_exp2}
\begin{split}
        \mathbb{E}[T_2] = \mathbb{E}\Bigg[ & \frac{D - A(c_k)}{\frac{1}{(C- c_k)}\int_{c_k}^C \alpha^{(m)} \psi(\nu) + \beta^{(m)} d\nu}\Bigg] \\ & \geq \frac{D - A(c_k)}{ \mathbb{E}\big[\frac{1}{(C- c_k)}\int_{c_k}^C \alpha^{(m)} \psi(\nu) + \beta^{(m)} d\nu\big]}
        \\ & = \frac{D - A(c_k)}{\big[ \mu_{\alpha} \cdot \frac{1}{(C- c_k)}\cdot \int_{c_k}^C \psi(\nu) d\nu + \mu_{\beta} \big]}
\end{split}
\end{equation}

Taking $(C - c_k) \to \infty$ on both sides and utilizing Proposition 2 in the final expression of Equation \ref{proof_lemma1_exp2} to obtain,
\begin{equation}\label{proof_lemma1_exp3}
    \begin{split}
        \lim_{(C - c_k) \to \infty} & \mathbb{E}[T_2] \\ &\geq  \frac{D - A(c_k)}{\lim_{(C - c_k) \to \infty} \big[\mu_{\alpha} \cdot \frac{1}{(C- c_k)}\cdot \int_{c_k}^C \psi(\nu) d\nu + \mu_{\beta}\big]} \\ &= \frac{D - A(c_k)}{\big[ \mu_{\alpha} \cdot \sum_{i \in S} (\pi_i \cdot \{\psi(.) = i\}) + \mu_{\beta}\big]} \\  & = \mathbb{E}[T_1] \\
        \text{or, \hspace{0.05cm}} & \lim_{(C - c_k) \to \infty} \mathbb{E}[T_2] \geq \mathbb{E}[T_1]
    \end{split}
\end{equation}
Using Equations \ref{proof_lemma1_exp1} and \ref{proof_lemma1_exp3}, we arrive at Equation \ref{lemma_Equation}.
\end{proof}  

\begin{remark}
     Lemma 1 shows that the expected lifetime computed using an effective degradation rate gives the conservative lower bound of the lifetime computed using Monte Carlo simulations in an almost sure sense. Note that, this is true if only a large number of sample paths (i.e., $M \to \infty$) are generated by simulations. Furthermore, $(C - c_k) \to \infty$ makes Lemma 1 true only for a long enough prediction horizon.
\end{remark}

\section{Simulation Studies}\label{Exper}

\subsection{Simulating a Task Planner}\label{TaskPlanner}
We simulate a task planner to generate a schedule of tasks with dynamic severity levels. In our study, the severity of a task is characterized by the weight of the payload being manipulated by the robot. One task is performed by the robot per operational cycle as discussed previously in Section \ref{modeling}. The inspection epochs are placed after 50 operational cycles. The position accuracy of the robot is observed only at those inspection epochs. 

In our experimental analysis, the most frequent task scheduled between two consecutive inspection epochs is modeled as a two-state CTMC. The first state of the CTMC corresponds to lifting a payload of 1 kg, the second corresponds to a 5 kg payload. The parameters of the CTMC are assumed to be deterministic for the purpose of the experiments. Table \ref{CTMC_table} mentions the parameters of the CTMC.  

We utilize two different types of physics-based simulators that represent -- a) planar and, b) spatial fleet of robotic manipulators. Our simulation included collecting task severity and position accuracy data from 25 robots of each fleet. These data were collected from the robot across a span of 300 inspection epochs, only after the onset of degradation in those robots. The robots fail (cross the threshold $D$) randomly at some cycle inside the span of those 300 epochs. 
\begin{table}
\centering
\caption{\label{CTMC_table}Parameters of the CTMC that represents the task planner}
\begin{tabular}{lc}
\toprule
\textbf{Parameter} & \textbf{Value}\\
\midrule
Q matrix & $\begin{pmatrix}
-0.005 & 0.005\\
0.005 & -0.005
\end{pmatrix}$ \\
Stationary distribution $\pi$ & $\begin{bmatrix}
0.5 & 0.5
\end{bmatrix}$ \\
\bottomrule
\end{tabular}
\vspace{-0.5cm}
\end{table}

\subsection{Simulating Degradation in Robotic Manipulators}
In this Section, we delineate the similarities between the planar and spatial simulators and, then describe the procedure for simulating degradation signals. Since the robotic simulations are an idealized representation of parallel long-lasting robot systems, there are many implicit common variables in both planar and spatial cases. For example, in either case, the simulation is a numerical solver for the forward dynamics of an open chain, rigid, multi-body system. This refers to solving Equation \ref{forward_dynamics}, formulated by \cite{lynch_modern_2017}, where $\theta$ are joint angles, $\dot{\theta}$ and $\ddot{\theta}$ are the first and second-time derivatives of the joint angles, $M$ is the mass matrix of the system, $\tau$ are applied torques at the joints, $C$ is the Coriolis matrix, and $g$ are gravitational forces. 
    \begin{equation}\label{forward_dynamics}
        \ddot{\theta} = M^{-1}(\theta)(\tau(t) - C(\theta,\dot{\theta})\dot\theta - g(\theta))
    \end{equation}
The control system is an integral part of simulating robotic motions in both planar and spatial cases. The applied torque $\tau$ in Equation \ref{forward_dynamics} necessitates the use of a closed-loop feedback controller at discrete times during each cycle. This controller computes the difference between the current and desired configuration of the robot system and commands torques to be applied at each joint at the given time. The controller torque signal must be able to drive power electronics that generate the required voltage to move the motors. Practically, there can be some amount of sensor imprecision, noise, and latency in this process but we assume it is negligible in simulation.

Besides, the aforementioned simulation variables, there are also two exogenous variables -- a) degradation and, b) task severity that are common for both cases. Robot joint damping (or viscous friction coefficient) is chosen as the degradation parameter in our experiments. This kind of degradation can be attributed to mechanical degradation like loss of lubrication in the gearbox or slop causing rubbing between joint components. In Equation \ref{friction}, $\tau_i$ is an additive torque on joint $i$, $\eta_i$ is the viscous friction coefficient for a given cycle at joint $i$ and $\dot{\theta}$ is the angular velocity.  
    \begin{equation}\label{friction}
        \tau_{i}(t) = -\eta_{i} * \dot{\theta}_{i}(t) 
    \end{equation}
We simulate degradation at two different joints of the robotic manipulators for both planar and spatial robots. The rate of change of the viscous friction coefficient (degradation) for each joint is a linear function of the weight of the payload lifted by the robot. Note that even if the degradation were present at the joint level, we observe only its manifestation on the position accuracy at the end-effector level. Figure \ref{fig:AccuracyPlot} shows the robot accuracy signals for five different robots from both types of robots. In either case, these signals are the measured position accuracy of the end-effector after the completion of a task cycle for that robot. 

The payload, the second of the exogenous variables, is represented differently in each of the experimental cases. In the planar case, it is attached to the end-effector, and in the spatial case it is grasped by the end-effector. The simulation does not consider closed-loop control of the payload itself, but only the end-effector. So the payload is not guaranteed to be in the same relative position as the end-effector over a given cycle and the payload jostling may feed back into the movement of the entire robot arm. This situation's effect on the robot's accuracy is assumed to be negligible.

Despite similarities in their simulation, there are some modeling differences between the planar and spatial robotic simulators. For example, the solution methods of Equation \ref{forward_dynamics} differ for both planar and spatial cases owing to differences in their complexities. In the next two subsections, we provide a detailed discussion of these simulators.

\subsection{Case Study 1: Planar Robotic Simulator}
\begin{figure}
    \centering
    \includegraphics[scale=0.25]{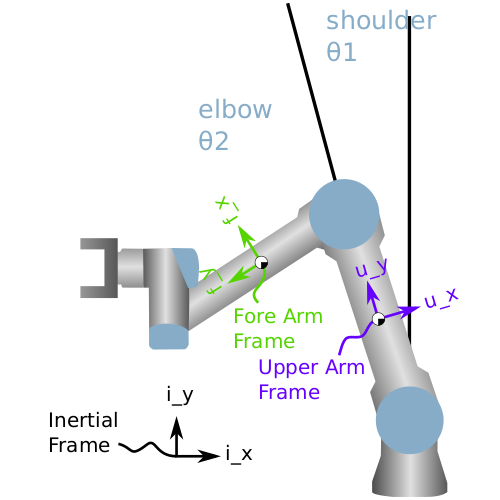}
    \caption{This is a schematic of the planar representation of the UR5e robot with rigid bodies labeled. The kinematic and inertial properties of the robot are accurate with the unstudied joints locked. The payload is rigidly attached to the end of the forearm link (not pictured).}
    \label{fig:ur5e}
    \vspace{-0.5cm}
\end{figure}
PyDy \cite{pydy2013} was chosen as the framework for building the planar robot simulation as it uses a symbolic math backend that expresses equations of motion in a human-readable format. The kino-dynamic model of the planar robot was built using Kane's method \cite{kane1983}: a process for defining equations of motion with forces in different frames acting on rigid bodies with constraints. In this case, the two joints of the robot represented two rigid bodies with external forces from gravity and the payload. Additionally, there are torques that depend of the controller outputs and degradation. 

The parameterization of the model follows the published specifications of the UR5e robot (Table \ref{ur5eparam}), with all degrees of freedom locked except the joints of interest. The controller consists of a gravity compensator and a linear-quadratic regulator (LQR). The LQR method produces an optimal state-feedback controller given a cost function specifying the relative importance of state deviation and minimizing effort (torque). The resulting controller acts at a defined frequency to minimize the difference between the observed and desired configuration of the robot. The costs of the minimization function are also tuned to achieve a reasonable response time with torques well below the stated maximums for the UR5e robot. 
Given the robot equations of motion with parameters from Table \ref{ur5eparam}, gravity compensator, and LQR controller which can be shown to be stable (see Appendix \ref{Stability}), the system was integrated using the LSODA \cite{Petzold1983} method from SciPy. This method switches between stiff and nonstiff integration methods which are particularly useful for systems in which small changes can induce large variability or instabilities such as the nonlinear case being explored. We use a 0.005-second integration time step and a 10-second cycle time for each motion. A schematic of the planar robot is given in Figure \ref{fig:ur5e}. {For the planar robot, a task is defined as a complete pick-and-place motion between two fixed joint configurations. Let the joint angles be defined as $\theta_1$ and $\theta_2$, which correspond to the upper arm and forearm, respectively. In each task, the robot is commanded to move from the joint configuration of $(\theta_1, \theta_2) = (10^{\circ}, 90^{\circ})$ to $(\theta_1, \theta_2) = (90^{\circ}, 20^{\circ})$.}
\begin{table}
    \centering
    \caption{UR5e Physical Parameters}
    \setlength{\tabcolsep}{0.5em}
    \begin{tabular}{*4c}
    \toprule
        \textbf{Parameter}  & \textbf{Value} & \textbf{Parameter}  & \textbf{Value}  \\ 
    \midrule
    Upper Arm Mass       & 8.393 kg & Fore Arm Mass        & 2.275 kg \\
    Upper Arm Radius     & 0.054 m  & Fore Arm Radius      & 0.060 m  \\
    Upper Arm Length     & 0.425 m  & Fore Arm Length      & 0.392 m  \\
    Upper Arm COM Length & 0.213 m  & Fore Arm COM Length  & 0.120 m \\
    \bottomrule
    \end{tabular}
    \label{ur5eparam}
    \vspace{-0.1cm}
\end{table}

\subsection{Case Study 2: Spatial Robotic Simulator}
 The simulation of a spatial robot with 7 degrees of freedom (DOF) was built with Robosuite\footnotemark[1] \cite{robosuite2020}, a toolbox for robotic manipulation studies using MuJoCo\footnotemark[2] (``Multi-Joint dynamics with Contact"), a general-purpose physics engine. In this study, it is especially useful because its actuation model includes a damping parameter as specified in equation \ref{friction}. 
\begin{figure}
    \centering
    \includegraphics[scale=0.15]{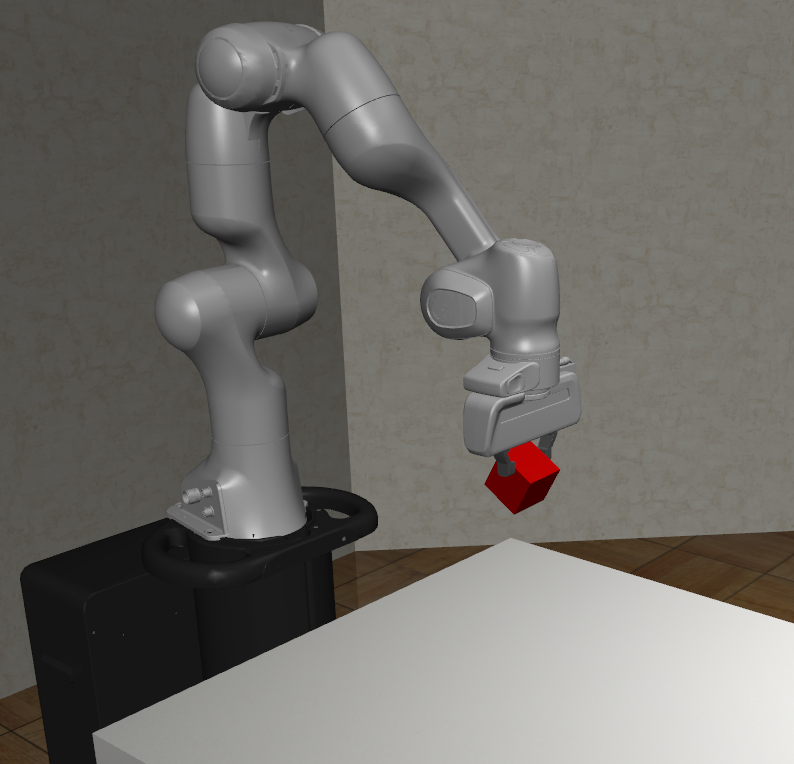}
    \caption{This 3D rendering of the Panda robotic arm depicts the final configuration of the arm for the commanded end-effector position with the payload. The payload was initially grasped from the table's surface. The payload is pitched because it slid while in transit.}
    \label{fig:panda}
     \vspace{-0.5cm}
\end{figure}
We utilize the Panda robot\footnotemark[3] model by Franka Emika\footnotemark[4] in this case study. This robot is a standard model from Robosuite with kinematic and inertial parameters from the real robot. The Panda robot grasps and moves a payload cube to the desired position in our case study. The 7 DOF of the Panda robot enables it to reach the desired end-effector position more easily because there is a continuum of joint configurations that give a valid end-effector pose. An image of the rendering of the 7 DOF panda robot is shown in Figure \ref{fig:panda}. 

An operational space controller is used, which commands torques to move the end-effector to the desired position with low kinematic energy. An end-effector trajectory is linearly interpolated and discretized from the goal pose to the current pose, and a desired wrench of the end-effector is computed at each point. A wrench is a ``spatial force" or moment and force expressed in a reference frame. The wrench is then propagated to the joints, given the Jacobian of the current joint configuration. The torque at each joint is clipped to the reported maximum rating, limiting the response of the robot upon reaching high degradation values. {For the spatial robot, a task is defined as a complete pick-and-place operation along a prescribed spatial trajectory segment. The robot begins a cycle in a joint configuration with the end effector at the cube with the grips open. Then, a trajectory is assembled which commands end-effector poses at certain times as listed in Table} \ref{trajectory}. {This controller has been shown to be stable (see Appendix} \ref{Stability}). {Each cycle is simulated for 5 seconds, enough time for the robot arm to reach the final position.}
\footnotetext[1]{Robosuite: \url{robosuite.ai}} \footnotetext[2]{MujoCo: \url{mujoco.org}} 
\footnotetext[3]{Panda robot: \url{robosuite.ai/docs/modules/robots.html}} \footnotetext[4]{Franka Emika: \url{franka.de}}
\begin{table}
    \centering
    \caption{Prescribed end-effector actions for spatial robot}
    \begin{tabular}{*2c}
    \toprule
    \textbf{End-Effector Pose}  & \textbf{Movement Start Time (s)}  \\ 
    \midrule
    Above cube         & 0                        \\
    Lower to cube      & 1                        \\
    Close gripper      & 2                        \\
    Move to the final pose & 3               \\
    \bottomrule
    \end{tabular}
    \label{trajectory}
    \vspace{-0.5cm}
\end{table}

\section{Results on Robot Lifetime Prediction}\label{results}

In either robot type, we first split the data into training and testing sets containing 24 and 1 robot, respectively. {Following the Bayesian prior estimation procedure described in \cite{Gebraeel2005}, we derive the prior mean and variance of the accuracy model parameters $(\alpha, \beta)$ from a training set of 24 robots. For each training robot, we estimate $\alpha$ and $\beta$ via least-squares regression by fitting Eq.~(2) to its observed data. This yields 24 paired estimates $(\hat{\alpha}_j, \hat{\beta}_j)$, $j=1,\dots,24$. The sample mean and variance of $\{\hat{\alpha}_j\}$ of the test robot are taken as the prior parameters ($\mu_1$, $\sigma_1^2$) for $\alpha$. Similarly, the sample mean and variance of $\{\hat{\beta}_j\}$ of the test robot define the prior parameters ($\mu_2$, $\sigma_2^2$) for $\beta$. This is repeated until we compute the prior parameters for all 25 robots (of each type).} 

The number of cycles after which the robot's accuracy crosses the fixed threshold (Section \ref{distribution}) is called the \textit{true lifetime} of the robot. In our case studies, the thresholds were specified to be 0.27 meters and 0.021 meters for the planar and spatial robots respectively. 
\begin{remark}\label{threshold_remark}
The failure threshold in prognostic modeling is often set by subject matter experts (SMEs). In our study, we determine it based on observed data, specifically as the maximum position accuracy crossed by all the 25 robotic manipulators. This criterion yielded different thresholds for planar and spatial robots, reflecting differences in their controllers.
\end{remark}
\begin{figure}
    \centering
  \subfloat[Planar \label{1a}]{%
       \includegraphics[width=0.5\columnwidth]{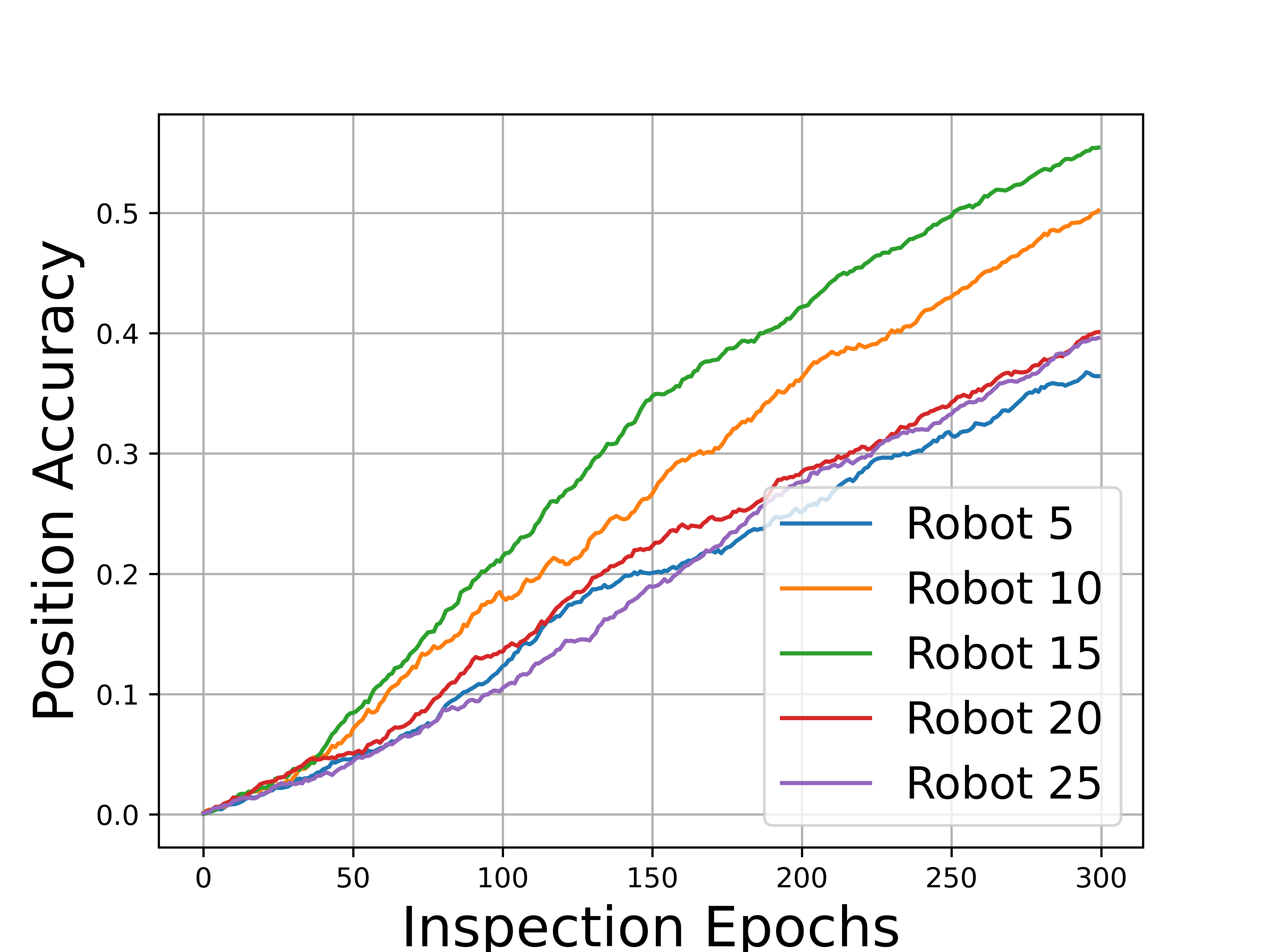}}
    \subfloat[Spatial\label{1a}]{%
       \includegraphics[width=0.5\columnwidth]{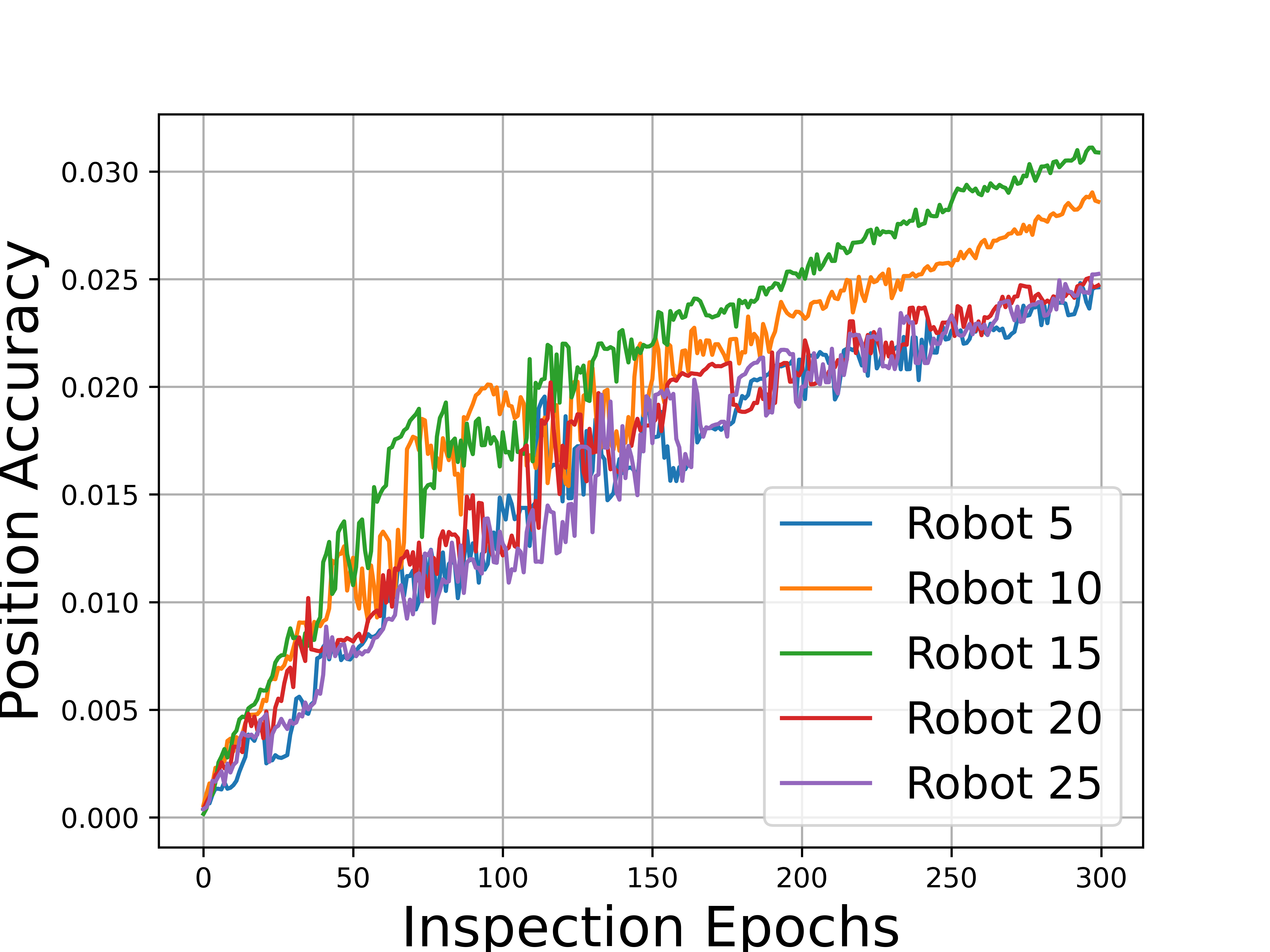}}
  \caption{Position accuracy of five (a) planar, and (b) spatial robots}
  \label{fig:AccuracyPlot} 
   \vspace{-0.5cm}
\end{figure}

\subsection{Numerical Analysis}
 In this section, we discuss experimental methods to validate the normality assumptions of accuracy model parameters $\alpha$ and $\beta$, and the Brownian noise assumption of the position accuracy model. We further provide numerical estimates of these parameters. 

Given the task planner and the robot accuracy data, least square estimates for $\alpha$ and $\beta$ were obtained for each robot. We performed the Shapiro-Wilks normality test on $\alpha$ and $\beta$ separately and their p-values were computed to be 0.89 and 0.59 respectively for the planar case and 0.31 and 0.62 respectively for the spatial case. Therefore, the test fails to reject the null hypothesis that $\alpha$ and $\beta$ are normally distributed in either fleet. After obtaining the estimate for the prior distribution of $\alpha$ and $\beta$ from the training robots, we perform Bayesian updating of these parameters at 4 different inspection epochs specifically at \{30\%, 50\%, 70\% \& 90\%\} of $L_f$ where $L_f$ refers to the true lifetime of the robot under consideration. After fitting the linear regression with the estimated $\alpha$ and $\beta$, we compute the error in our accuracy model by subtracting the accuracy data and the fitted model value for inspection epochs $c_i, \hspace{0.2cm} \forall i > 0$. The error increments are then computed for each pair of consecutive inspection epochs $c_{i-1}$ and $c_i$ for all $i > 0$, for all test robots. These error increments are expected to follow a normal distribution with mean 0 and standard deviation $\gamma \sqrt{(c_i - c_{i-1})}$ for each test robot. First, we found out that all these increments follow a normal distribution. 
In addition, the error increments were also found to have a random variation around 0 therefore, validating our Brownian error assumptions. Recall that we had 50 operational cycles between any two inspections in our experiments. Therefore, we divided the variance (across error increments) by 50 to obtain an estimate of $\gamma^2$ for that training robot. For any test robot, we obtained the mean of $\gamma$ across 24 training robots to serve as the deterministic $\gamma$.

\subsection{Remaining Lifetime Distribution}
\begin{figure*}
    \centering
    \subfloat[Planar fleet\label{1a}]{%
        \includegraphics[width=0.5\textwidth]{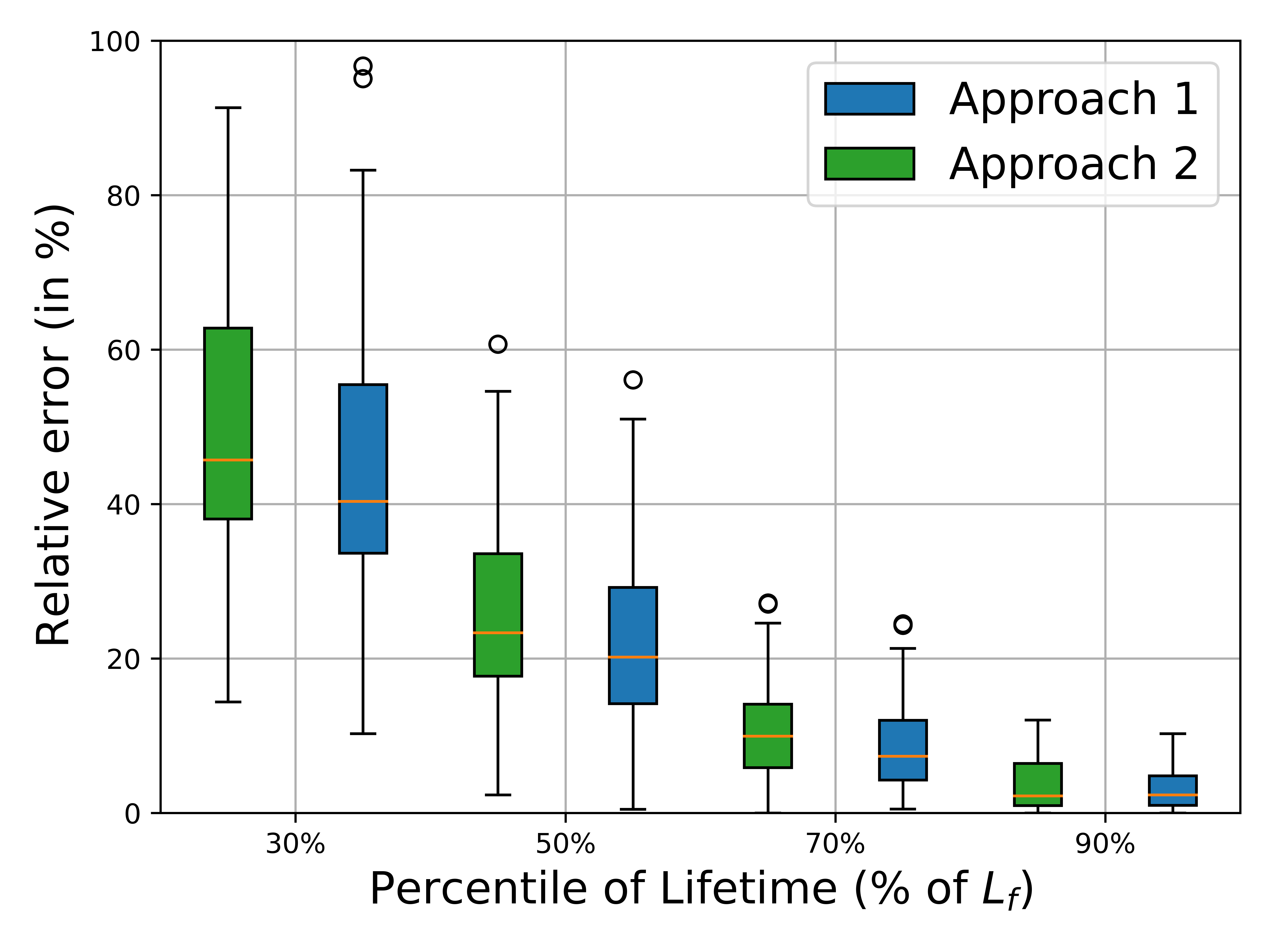}}
  \subfloat[Spatial fleet\label{1b}]{%
        \includegraphics[width=0.5\textwidth]{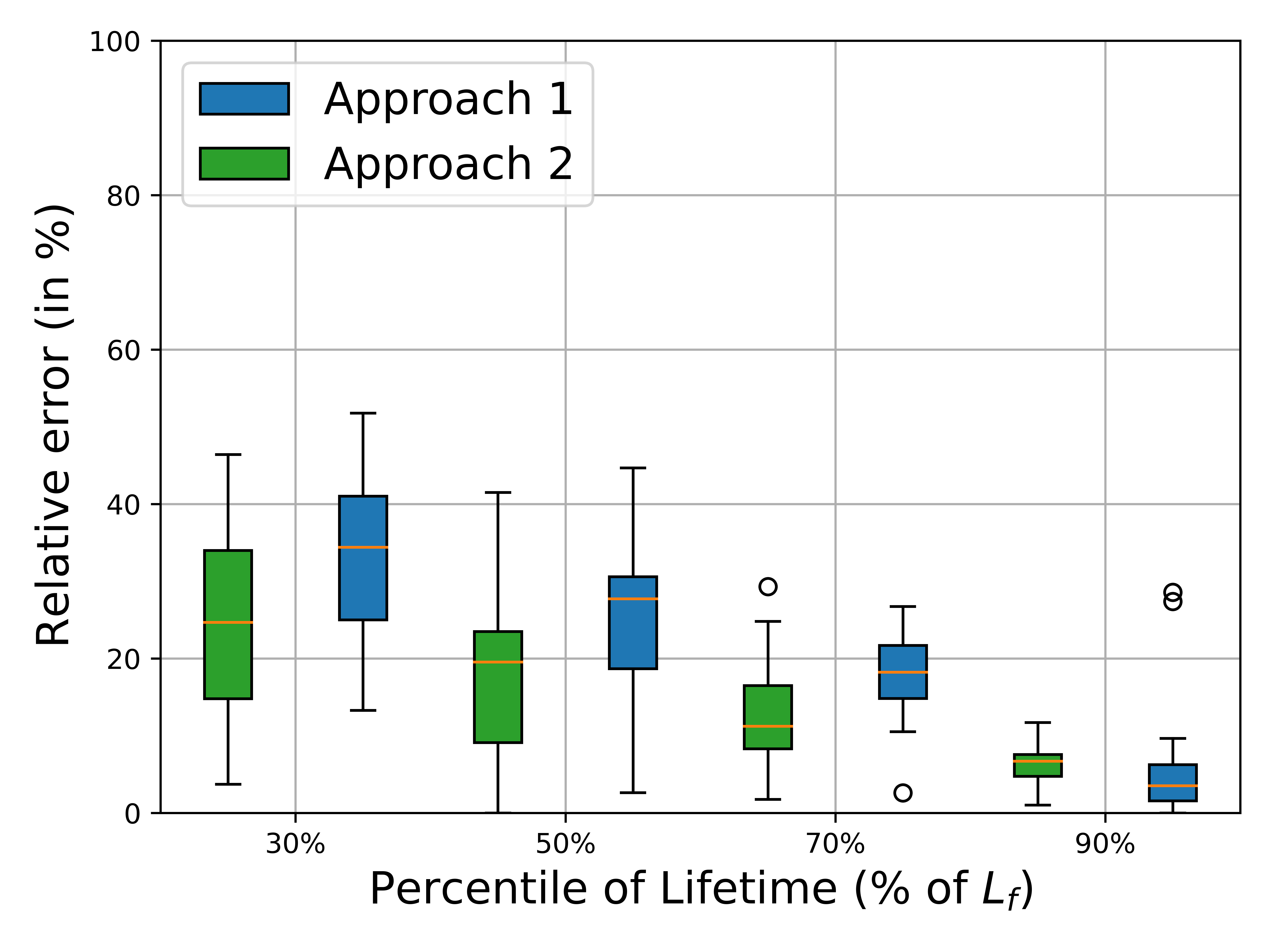}}
  \caption{Prediction errors using approaches 1 and 2 across all robots of (a) planar fleet and (b) spatial fleet}
  \label{fig:PredictionError_6DOF} 
\end{figure*}

Using the estimated values of the parameters, we compute the RLD of the robots at $30\%, 50\%, 70\%,$ \& $90\%$ of their lifetime. In our results, the predicted RUL of the robot under study at inspection epoch $c_k$ is considered to be the median of the obtained RLD $\hat{R}_k$ i.e., where  $P(\hat{R}_k | {\mathcal{A}_{c_k}}, {\mathcal{P}_{c_k}}) = 0.5$. To compute the prediction errors of our approaches, we obtain the \textit{predicted lifetime} of the robot from its remaining functional lifetime. Let $\hat{R}_k$ be the predicted RFL of the robot at inspection epoch $c_k$, then we compute predicted lifetime $\hat{L}_{f_i}$ using Equation \ref{predicted_lifetime}. 
\begin{equation}\label{predicted_lifetime}
    \hat{L}_{f_i} = \hat{R}_k + c_k
\end{equation}
Thus, if the true lifetime of the robot under study is given as $L_f$, then the prediction error is given by Equation \ref{prediction_error}. 
\begin{equation}\label{prediction_error}
    \text{Prediction error} = \frac{|\hat{L}_{f_i} - L_f|}{L_f} \times 100
\end{equation}
Figure \ref{fig:PredictionError_6DOF} shows the prediction error across robot types at each update point. The error decreases as more information is gathered closer to failure, with both approaches exhibiting similar errors across fleets.

\label{results:FLD}

\subsection{Computation Time}
Table \ref{table:comparison} compares the computational time for both approaches across the planar and spatial fleets, representing the total time required to train and test all 25 robots in each fleet. The experiments were conducted on a macOS M1 chip system with 16 GB of memory. 
Approach 1 significantly outperforms Approach 2 in computation time due to its closed-form RLD computation, whereas Approach 2 relies on a numerical method requiring the generation of a large number of sample paths. {In Approach 2, the computational effort is directly proportional to the number of sample paths used to approximate the RLD; higher numbers of paths yield better accuracy but require more computation time. In our experiments, we used $M = 1000$ sample paths, which provided a reasonable balance between computational cost and estimation accuracy, though this still results in substantially longer runtimes compared to the closed-form computation in Approach 1}
\begin{table}
    \centering
    \caption{Computational time (in seconds) for approaches 1 and 2}
    \begin{tabular}{*3c}
    \toprule
        \textbf{Fleet}  & \textbf{Approach-1}  & \textbf{Approach-2}\\ 
      \midrule

    \textbf{Planar Robots} & {5.49} & {11080.37}\\

    \textbf{Spatial Robots} & {5.09} & {9218.62}\\
\bottomrule
    \end{tabular}
    \label{table:comparison}
    \vspace{-0.2cm}
\end{table}
\label{compare}

\subsection{Performance Comparison}\label{baselines}
We compare our approaches with two widely used data-driven prognostic models from the literature. First, a Brownian motion-based position accuracy (degradation) model with a \textbf{fixed degradation rate}, independent of task severity as utilized in \cite{Gebraeel2005, Qibo2021, Elwany2009, Aivaliotis_Arkouli_Georgoulias_Makris_2021, Bian2012}. While simple and commonly adopted, this model underestimates the impact of task severity (operating conditions) on degradation. Thus, this comparison underscores the importance of incorporating task-dependent effects, as done in our approaches. 

Second, we evaluate a baseline where our Brownian motion-based position accuracy (degradation) model is employed with a \textbf{deterministic task severity} (operating condition) profile, i.e.,  $\psi$ known apriori and not modeled as a CTMC. These types of degradation models have been studied in \cite{Liao2013, Hong2015, Tamssaouet2021}. This assumes perfect foresight of future tasks, which is often unrealistic in real-world scenarios. Comparing against this deterministic model demonstrates the benefits of modeling task severity as a stochastic process governed by a CTMC. 

Table \ref{tab:ComparisonBaselines} presents the error in predicting the RUL across all 25 robots for both the planar and spatial fleets. The prediction errors for the two baselines — \textbf{(1)} Fixed Degradation Rate and \textbf{(2)} Deterministic Task Severity Profile — are shown in the last two columns. Our proposed Approach-1 and Approach-2 significantly outperform the Fixed Degradation Rate baseline, which exhibits large prediction errors due to its inability to account for task severity. Additionally, while the Deterministic Task Severity baseline performs comparably in most scenarios, its reliance on perfect foresight makes it an infeasible option. 
\begin{table}
\centering
\caption{Prediction error (in \%) for different prognostic models}
\setlength{\tabcolsep}{0.3em}
\begin{tabular}{llcccc}
\toprule
\textbf{Fleet} & \textbf{\% of $L_f$} & \multicolumn{2}{c}{\textit{\text{Our Approaches}}} & \multicolumn{2}{c}{\textit{\text{Baselines}}}\\
{} & {} & \textbf{Approach-1} & \textbf{Approach-2} & \textbf{Fixed Rate} & \textbf{Deter. Task} \\ \midrule
& 30\% & 40.5 $\pm$ 19.4 & 46.3 $\pm$ 20.2 & 228.5 $\pm$ 80.1 & 41.9 $\pm$ 18.8 \\ 
                        Planar & 50\% & 21.5 $\pm$ 12.2 & 25.1 $\pm$ 12.7 & 152.4 $\pm$ 43.8 & 26.0 $\pm$ 14.5 \\  
                        Robots & 70\% & 8.0 $\pm$ 5.4 & 9.7 $\pm$ 6.6 & 58.4 $\pm$ 20.9 & 10.5 $\pm$ 7.3 \\ 
                        & 90\% & 2.9 $\pm$ 2.5 & 3.6 $\pm$ 3.2 & 14.3 $\pm$ 8.5 & 3.2 $\pm$ 2.6 \\ \midrule
& 30\% & 33.2 $\pm$ 11.5 & 24.4 $\pm$ 12.7 & 19.6 $\pm$ 14.9 & 25.4 $\pm$ 11.9 \\ 
                        Spatial & 50\% & 25.0 $\pm$ 10.4 & 18.1 $\pm$ 10.4 & 16.3 $\pm$ 12.3 & 18.7 $\pm$ 10.5 \\ 
                         Robots & 70\% & 18.1 $\pm$ 4.3 & 12.1 $\pm$ 6.1 & 13.7 $\pm$ 5.7 & 13.0 $\pm$ 7.4 \\ 
                         & 90\% & 3.3 $\pm$ 2.6 & 6.4 $\pm$ 2.6 & 5.3 $\pm$ 6.3 & 6.3 $\pm$ 2.8 \\ \bottomrule
\end{tabular}
\label{tab:ComparisonBaselines}
\vspace{-0.25cm}
\end{table}

\section{Impact of Task Proportions on Robot Lifetime}\label{whatif}
The results shown in Section \ref{results} assumed that the robot was subjected to the same task proportion throughout its lifetime. These proportions were represented by the stationary distribution $\pi$ of the task severity CTMC (Table \ref{CTMC_table}). As noted in Remark \ref{remark_stationaryDistribution}, $\pi$ can be estimated empirically from historical task sequences, eliminating the need to compute the $Q$ matrix. Approach 1 thus gives a closed-form expression for practitioners to utilize \textbf{\textit{future task proportions as a tuning parameter}} for analyzing the robot's lifetime. In this section, we utilize this capability to analyze the changes in robot lifetime under several ``\textit{what-if future scenarios of task proportions}''.

\subsection{What-if Scenarios}
At each update point, the robot's lifetime is predicted for five different \textit{what-if scenarios} given by  $\pi = [1, 0], [0.75, 0.25], [0.5, 0.5], [0.25, 0.75],$ and $[0, 1]$ for all 25 robots in both types of robots. Here, $\pi = [x, 1-x]$ corresponds to $x \times 100 \%$ of 1 kg tasks and $(1-x) \times 100 \%$ of 5 kg tasks scheduled by the planner. 

\begin{figure*}
    \centering
    \subfloat[30\% $L_f$, Planar robot\label{1b}]{%
        \includegraphics[width=0.25\textwidth]{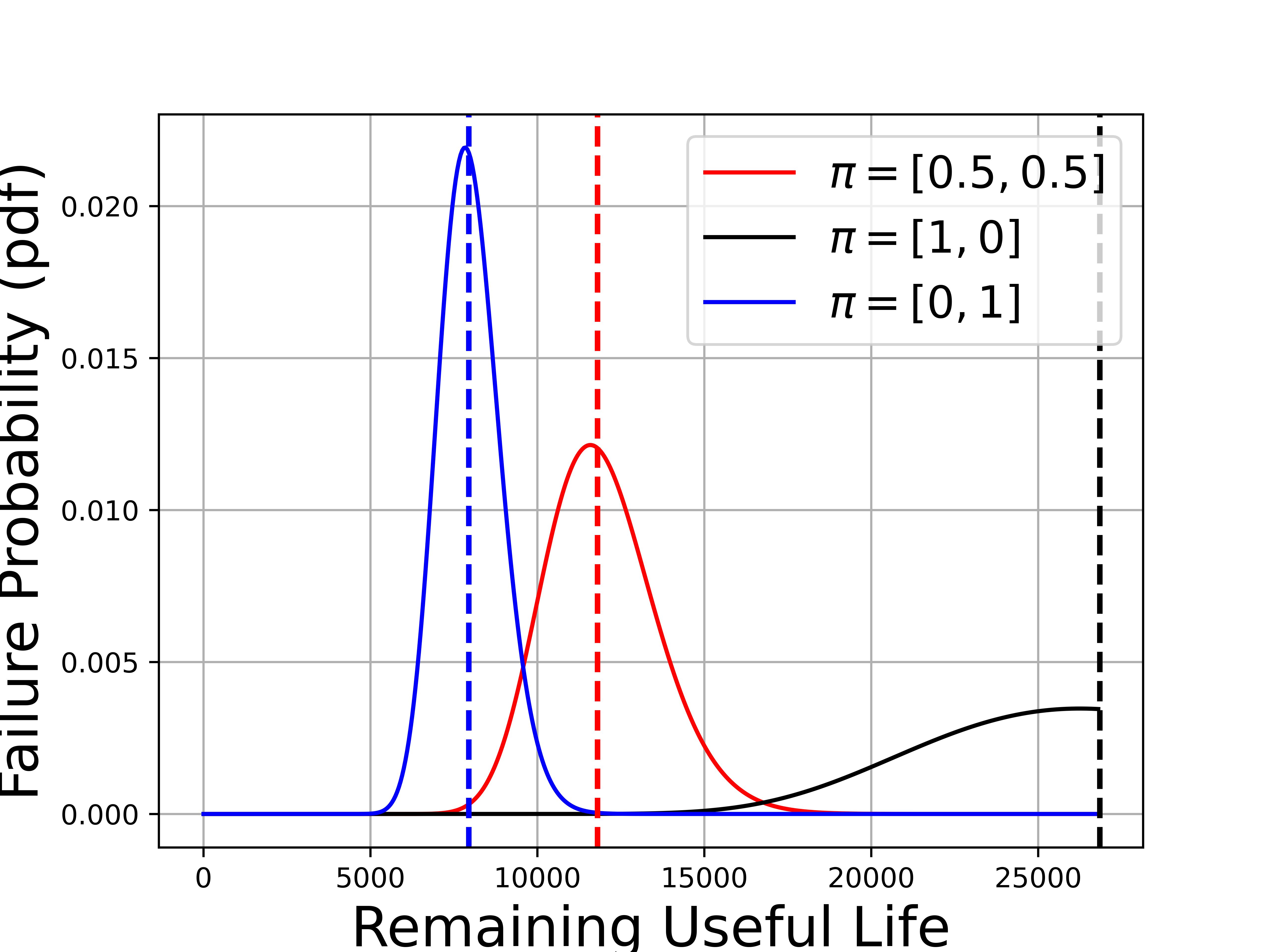}}
  \subfloat[50\% $L_f$, Planar robot\label{1d}]{%
        \includegraphics[width=0.25\textwidth]{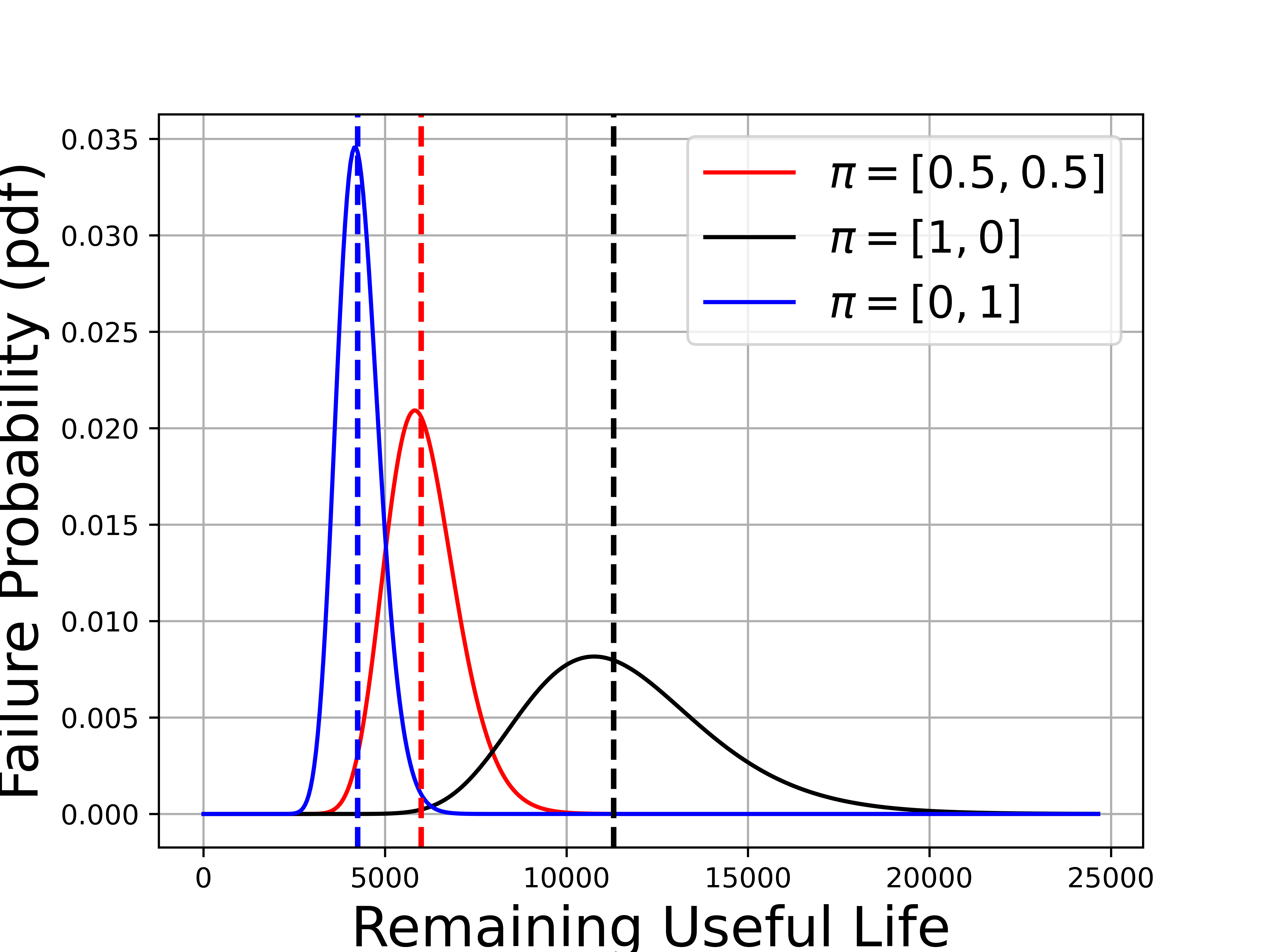}}
  \subfloat[70\% $L_f$, Planar robot\label{1d}]{%
        \includegraphics[width=0.25\textwidth]{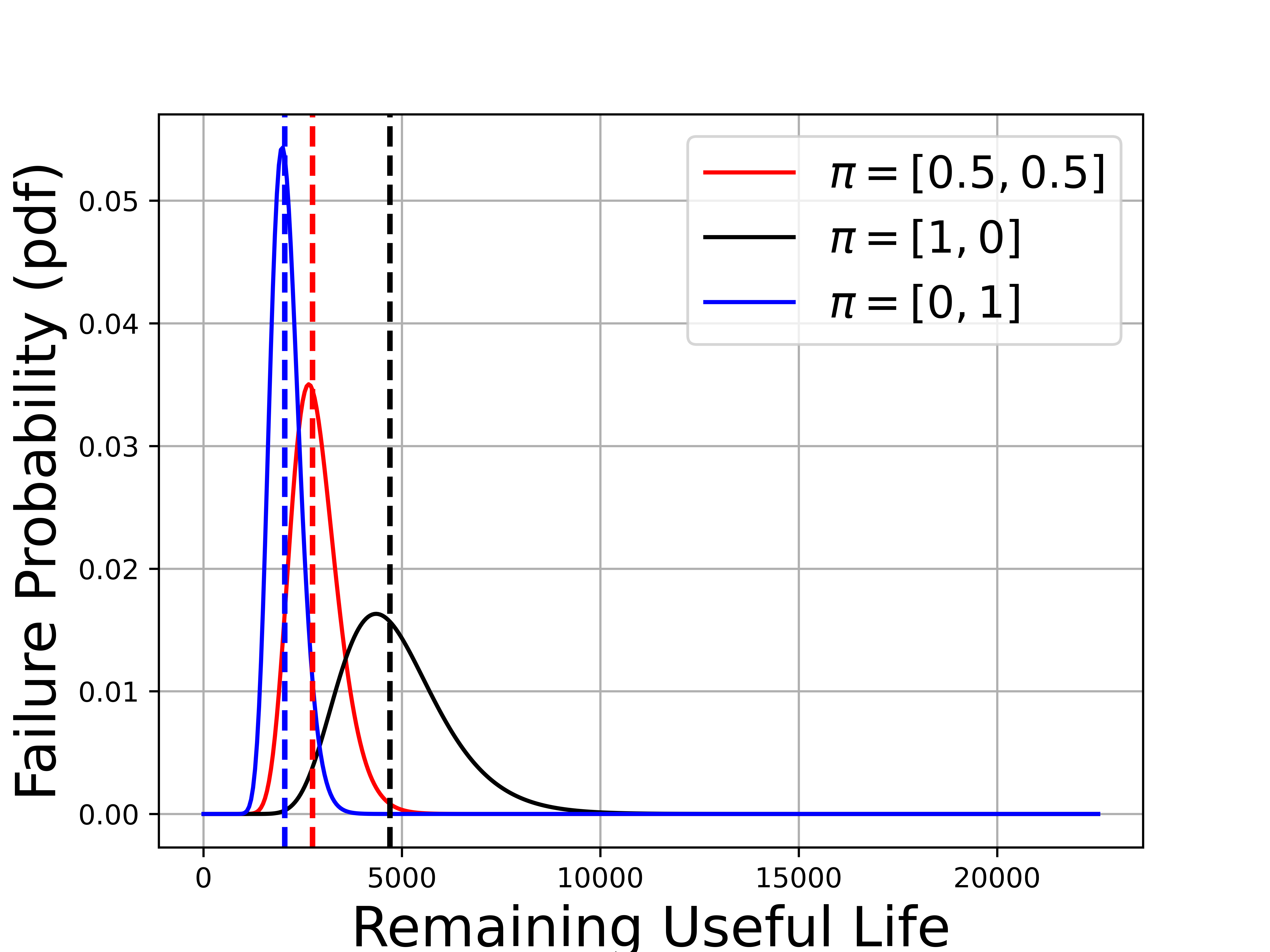}}
  \subfloat[90\% $L_f$, Planar robot\label{1d}]{%
        \includegraphics[width=0.25\textwidth]{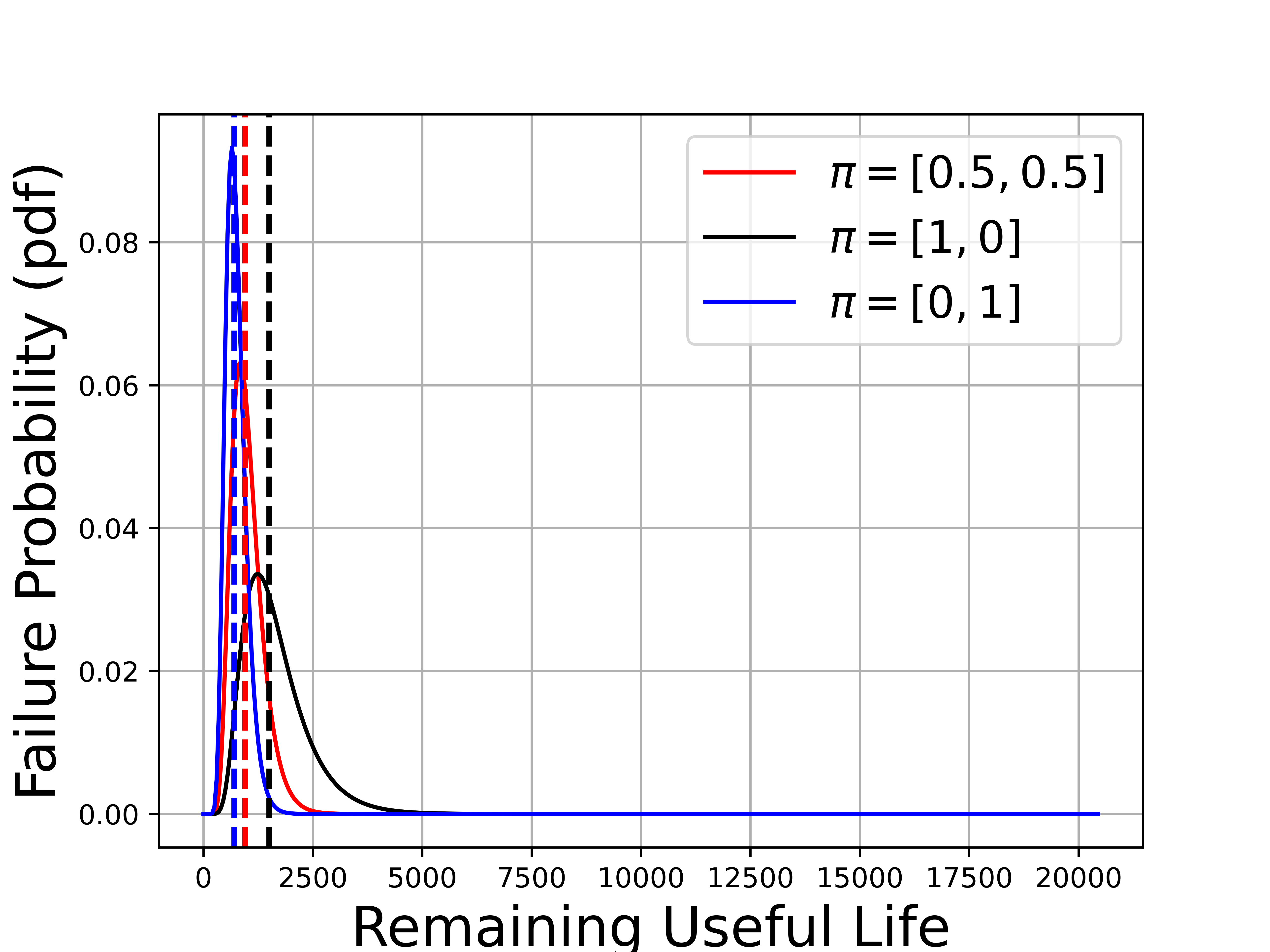}}
    \hfill
  \subfloat[30\% $L_f$, Spatial robot\label{1b}]{%
        \includegraphics[width=0.25\textwidth]{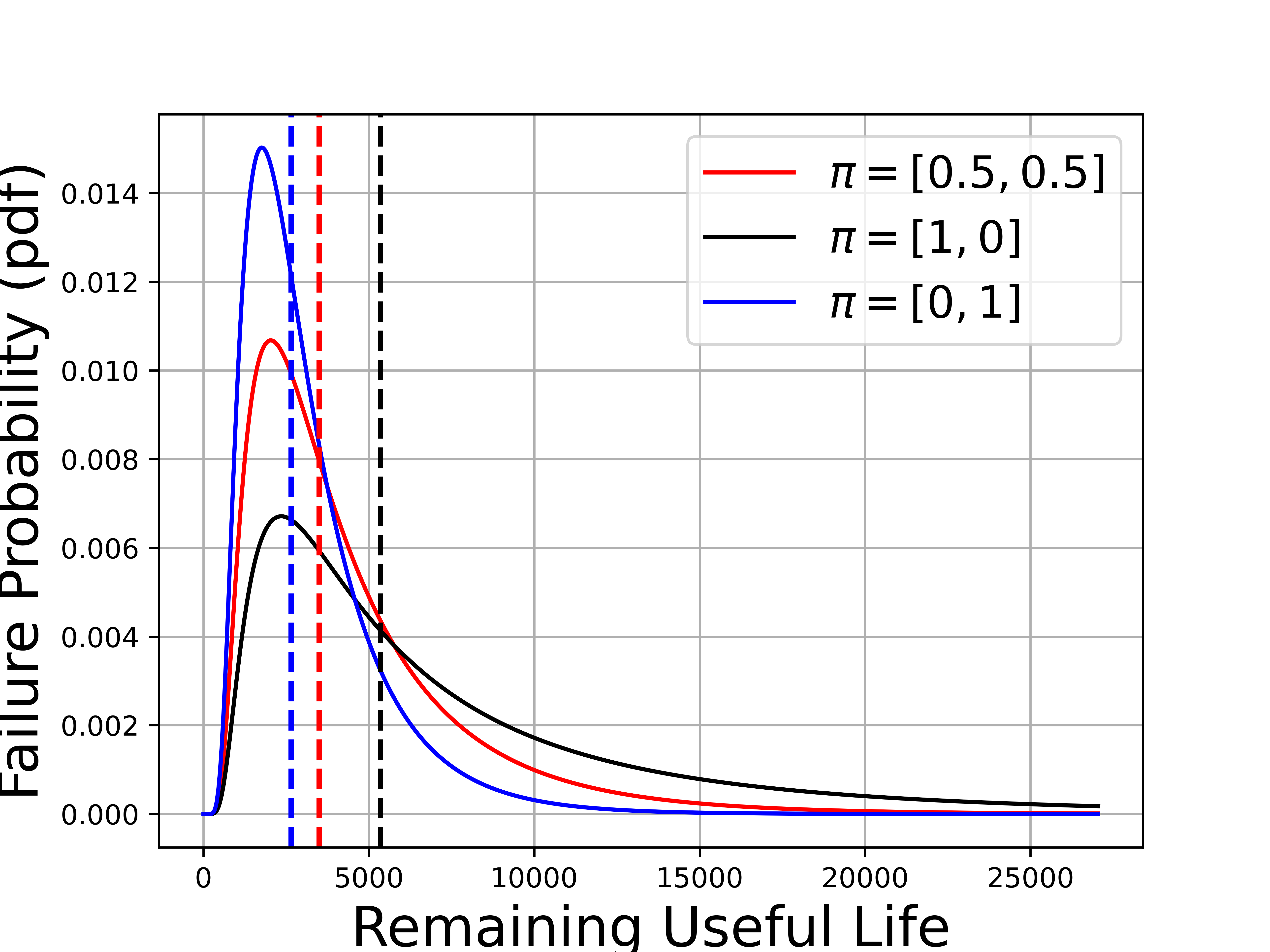}}
  \subfloat[50\% $L_f$, Spatial robot\label{1d}]{%
        \includegraphics[width=0.25\textwidth]{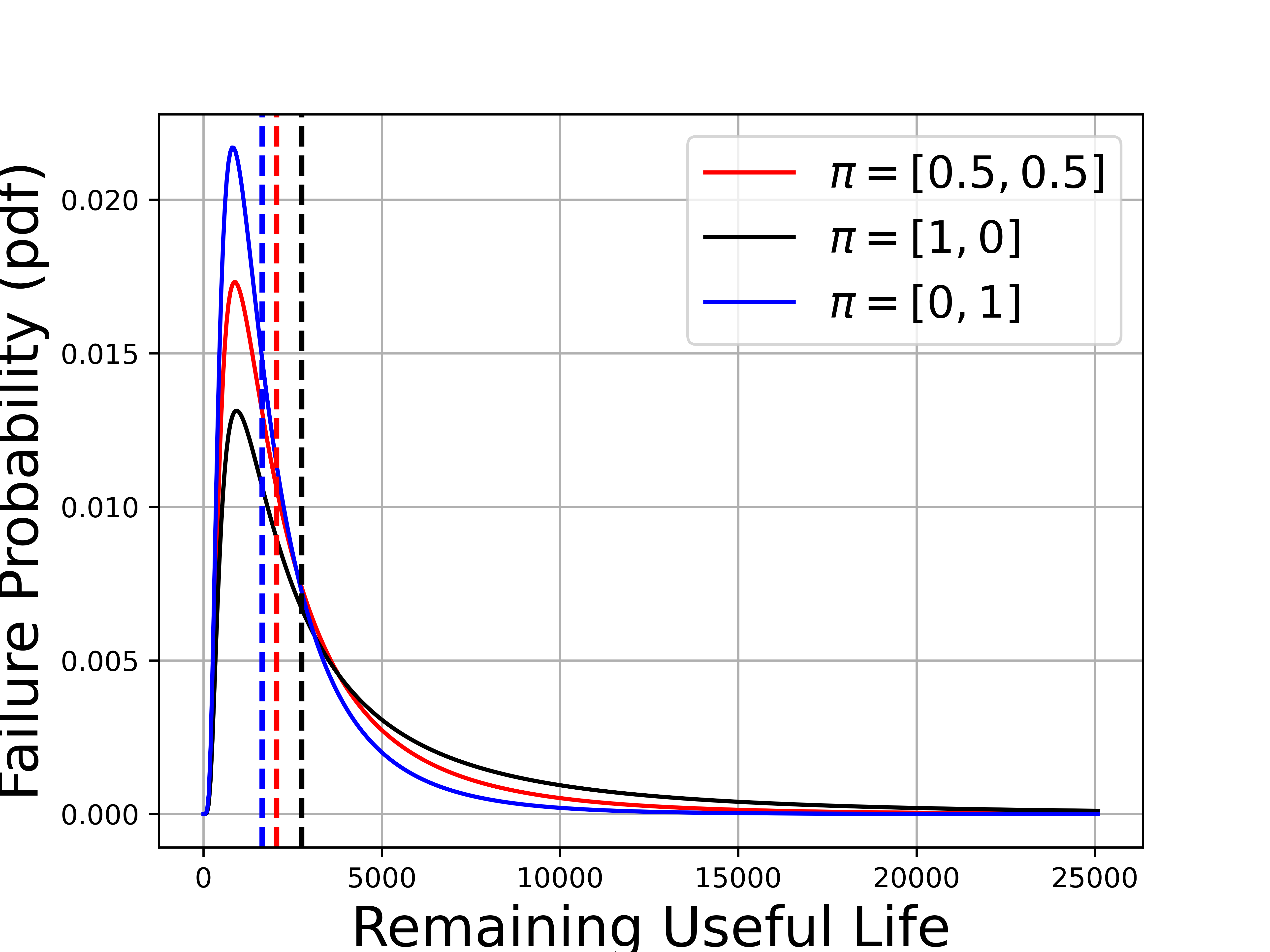}}
        \hfill
  \subfloat[70\% $L_f$, Spatial robot\label{1d}]{%
        \includegraphics[width=0.25\textwidth]{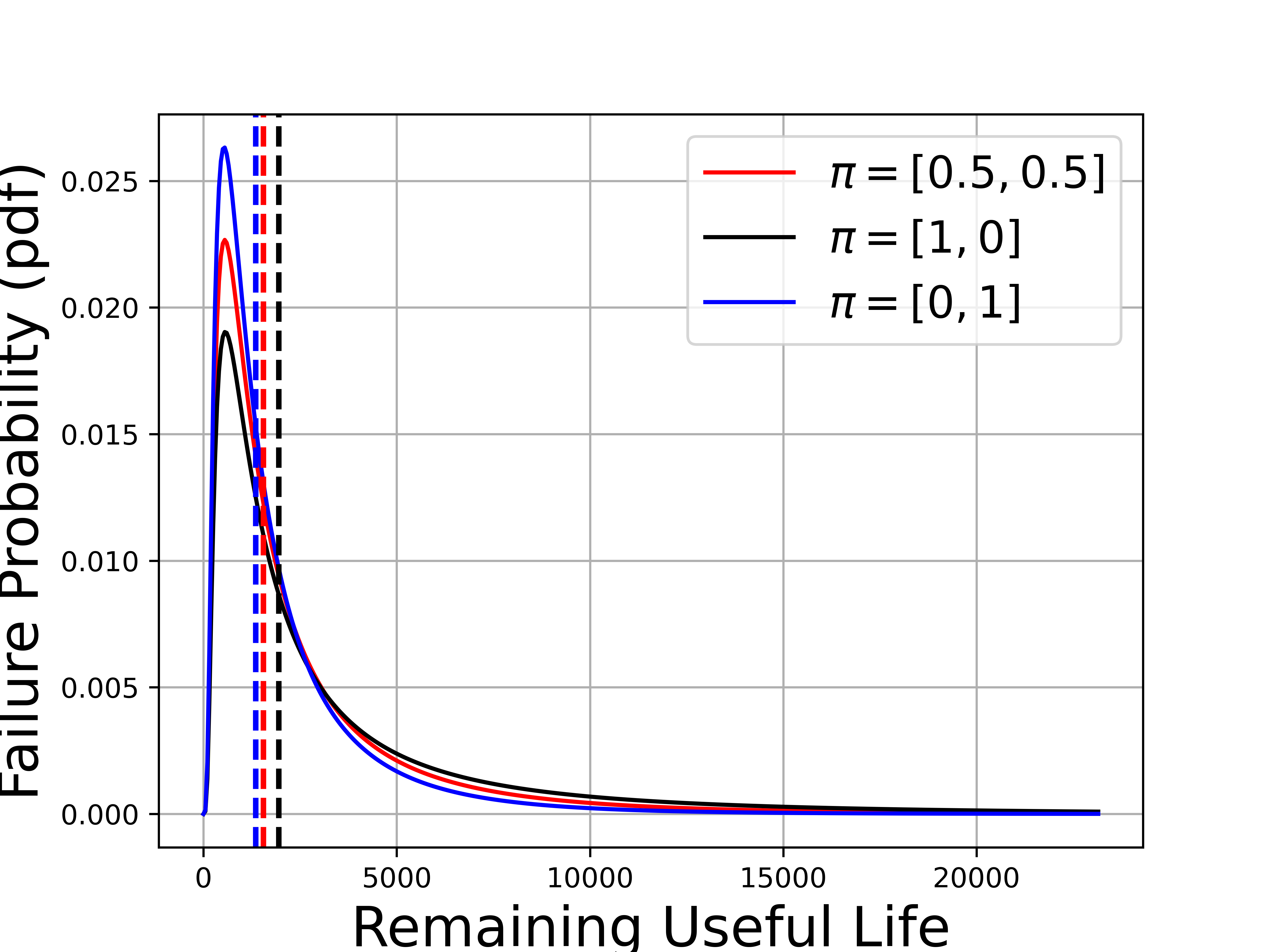}}
  \subfloat[90\% $L_f$, Spatial robot\label{1d}]{%
        \includegraphics[width=0.25\textwidth]{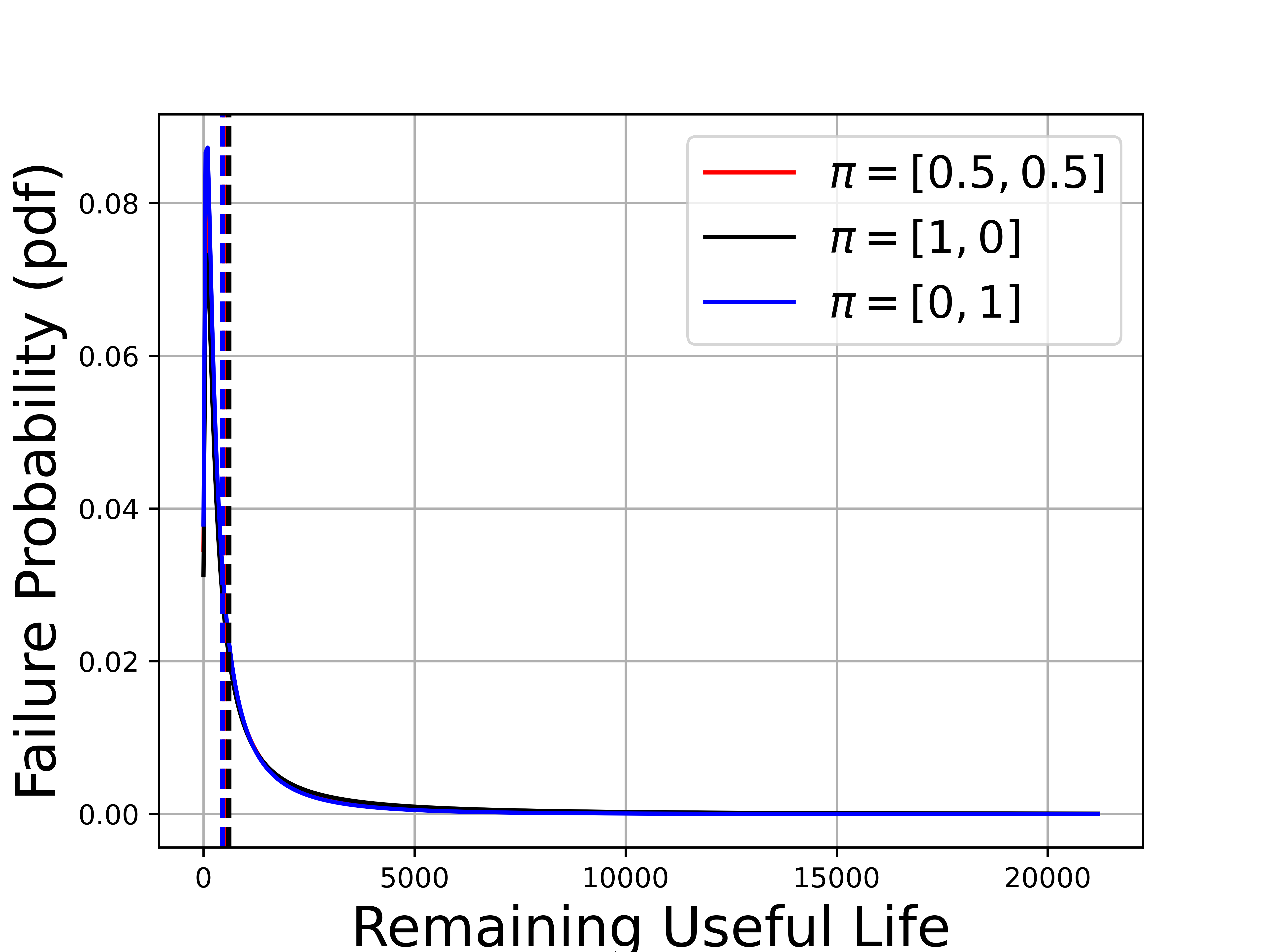}}
  \caption{The remaining lifetime distributions for a chosen planar and spatial robot, for 3 different task proportions. The RUL of the robots is given by the median of the distributions and denoted by the dotted lines. The color schema is same for the distribution and its median. }
  \label{fig:FLD_whatif_6DOF} 
\end{figure*}

Table \ref{table:spatialRobot_Whatif} states the predicted lifetime for a chosen robot from planar and spatial types. Figure \ref{fig:FLD_whatif_6DOF} shows the distributional shift for a planar and spatial robot at the current task proportion and two extremes, i.e., $\pi = [1, 0], [0.5, 0.5],$ and $[0, 1]$. The RUL in Figure \ref{fig:FLD_whatif_6DOF} is given in the number of operational cycles (i.e., $\hat{R}_k$ $\times$ number of operation cycles between two consecutive inspection epochs). 

In Table \ref{table:spatialRobot_Whatif}, for both planar and spatial cases, the lifetime of the robot decreases with an increase in future proportions of 5 kg tasks. It is also evident by shifts depicted by the distribution plots of either of the extreme scenarios in Figure \ref{fig:FLD_whatif_6DOF}. Furthermore, in Table \ref{table:spatialRobot_Whatif} for both types of robots, the impact of task proportions on the robot's predicted lifetime decreases as we go closer to failure (i.e., from 30\% to 90\%  of $L_f$). The plots also suggest that there is a distributional shift in the early stages of inspection, but as we approach failure, the differences in the distributions decrease. In Figure \ref{fig:FLD_whatif_6DOF}, we also observe that the magnitude of the distributional shift in the spatial robot is much lower than its planar counterpart. This might be due to the high noise present in the accuracy measurements for the spatial robot (Figure \ref{fig:AccuracyPlot}). 

\begin{table*}
\centering
\caption{Comparison of the predicted lifetime under different task proportions for a chosen robot from both planar and spatial fleet}
\setlength{\tabcolsep}{1.2em}
\begin{tabular}{*7c}
\toprule
{} & {} & \multicolumn{5}{c}{\textbf{Proportions of Future Tasks}}\\
\cmidrule[1pt](lr){3-7}
{}   & {}   & $\pi = [1, 0]$ & $\pi = [0.75, 0.25]$    & $\pi = [0.5, 0.5]$   & $\pi = [0.25, 0.75]$ & $\pi = [0, 1]$\\
\midrule
\textbf{Robot Fleet \& ID} &  \multicolumn{1}{c}{\textbf{Update points (as \% of $L_f$)}} & \multicolumn{5} {c}{\textbf{Predicted Lifetime (in number of operational cycles)}}\\
\midrule
{}   & 30\% & 30000	&20200	&14950	&12800	&11100 \\
Planar Robot 5   & 50\% &16600	&13300	&11300	&10350	&9550 \\
{}   & 70\% & 12100	&10950	&10150	&9800	&9450\\
{}   & 90\% & 11000	&10700	&10450	&10350	&10200  \\
\midrule
{}   & 30\% & 8250	&7150	&6400	&5950	&5550 \\
Spatial Robot 5   & 50\% &7600	&7200	&6900	&6700	&6500  \\
{}   & 70\% & 8750	&8550	&8350	&8250	&8150  \\
{}   & 90\% & 9350	&9300	&9250	&9200	&9200  \\
\bottomrule
\end{tabular}\label{table:spatialRobot_Whatif}
\end{table*}

\section{Limitations}
The proposed framework has several limitations. Approach 1 assumes an ergodic Markov chain, which may not represent real-world task schedulers and relies on long-run Markov chain behavior for accuracy, necessitating sufficiently large prediction horizons. While Approach 2 achieves greater accuracy with larger sample sizes, it is computationally expensive due to the need to generate a large number of sample paths. Additionally, the experimental validations were simplified using known CTMC parameters and a limited task severity space (e.g., {1kg, 5kg}) to reduce computational complexity. Physics-based robotic simulators were utilized in this study due to the high cost and logistical challenges of conducting real-robot failure experiments.

\section{Conclusion}
This paper presents a framework for predicting the lifetime of partially degraded robotic manipulators performing tasks of varying severity, characterized by payload weights. Degradation is measured by the position accuracy of the end-effector, with lifetime defined as the point where accuracy exceeds a predefined threshold. A periodic inspection scheme collects accuracy data, modeled as Brownian motion with drift dependent on task severity. A CTMC models task scheduling, with parameters periodically updated using Bayesian methods. We propose two approaches to estimate the robot's remaining lifetime distribution (RLD): a closed-form solution for computational efficiency and a simulation-based method. Both showed comparable predictive accuracy in simulated studies of planar and spatial manipulators. Approach 1's closed-form expression provides a practical tool for visualizing RUL under different task scenarios, revealing that higher proportions of severe tasks reduce lifetime.
\section*{Acknowledgement}
This effort is supported by NASA under grant number 80NSSC19K1052 as part of the NASA Space Technology Research Institute (STRI) Habitats Optimized for Missions of Exploration (HOME) ‘SmartHab’ Project. Any opinions, findings, conclusions, or recommendations expressed in this material are those of the authors and do not necessarily reflect the views of the National Aeronautics and Space Administration.
\appendices
\section{Proof of Proposition 1}\label{Proof_Prop1}
Substituting Equation \ref{acc_likelihood} in  Equation \ref{bayes_acc_2} and using the distribution of $\alpha$, $\beta$ we have,
\begin{equation}\label{posterior_accuracy_proof}  
\begin{split}
    \nu_A(\alpha, \beta |\mathcal{A}_{c_k}, \mathcal{PL}_{c_k}) & \propto \pi_A(\alpha)\pi_A(\beta) \mathit{f}(\mathcal{A}_{c_k} | \mathcal{PL}_{c_k}, \alpha, \beta) \\\propto exp\Bigg( &-\frac{(\alpha - \mu_1)^2}{2\sigma_1^2} -\frac{(\beta - \mu_2)^2}{2\sigma_2^2}  \\ & - \frac{1}{2} \sum_{i = 1}^k  \frac{\big(A_i -  [\alpha \psi_i + \beta (c_i - c_{i-1})]\big)^2}{\gamma^2(c_i - c_{i-1})}\Bigg)
\end{split}
\end{equation}
By comparing the coefficients of $\alpha^2$, $\beta^2$, $\alpha\beta$, $\alpha$ and $\beta$ in Equation \ref{posterior_accuracy_proof} and the posterior distribution of a bivariate normal with the same parameters, we obtain the expressions for $\mu_\alpha, \mu_\beta, \sigma_\alpha, \sigma_\beta \hspace{0.1cm} \text{and} \hspace{0.1cm} \rho$ given in proposition 1
\section{Proof of Proposition 2}\label{Proof_Prop2}
We assumed that the CTMC is ergodic. Therefore, following the ergodic theorem of Markov chain (\cite{serfozo2009}),  we have,
\begin{equation}\label{ergodic}
    \lim_{c \to \infty} \frac{1}{c} \cdot \int_{0}^{c} f(\psi(\nu)) \, d\nu = \sum_{i \in S} \pi_i \times f(\psi(\nu) = i)
\end{equation}
where, $\psi(.)$ is the ergodic CTMC with state space $S$, $\pi$ is the stationary distribution of the CTMC . 

Considering $f(.)$ in eq \ref{ergodic} to be the identity function (i.e., $f(\{\psi(.) = i\} ) = \{\psi(.) = i\} $) we obtain,  
\begin{equation}
    \lim_{c \to \infty} \frac{1}{c} \cdot \int_{0}^{c} \psi(\nu) d\nu = \sum_{i \in S} \pi_i \times \{\psi(.) = i\} 
\end{equation}
\section{Notes on Stability Analysis}\label{Stability}
\begin{itemize}
    \item \textbf{LQR Controller:} For the \textit{planar robot} case, we utilized an LQR controller for its simplicity in defining the relative importance of state error and control effort. The non-linear two-link manipulator system was linearized around its operating point, following the methodology outlined in Section 8.2 of \cite{underactuated}. By ensuring that the design matrices $R$ and $Q$ satisfy the conditions of being positive definite and positive semi-definite, respectively, our implementation adheres to the stability properties described in \cite{underactuated}, confirming the system's stability.
    \item \textbf{Operational Space Controller:} The operational space controller used for the \textit{spatial robot} in our experiments is based on the formulation provided by Khatib \cite{khatib}. Notably, the Lyapunov stability of the controller, as described in \cite{khatib}, is ensured by the dissipative property outlined in Equation 59 of \cite{khatib}. Since our implementation incorporates such a dissipative term in the controller implemented in \cite{robosuite}, therefore, the system is stable, provided that the redundant degrees of freedom are also stabilized. As noted in \cite{khatib}, applying a dissipative force within the null space stabilizes the redundant degrees of freedom, a feature included in our implementation.
\end{itemize}

\bibliographystyle{IEEEtran}
\bibliography{ref.bib}

\section*{Biography Section}
\begin{IEEEbiography}[{\includegraphics[width=1in,height=1.15in,clip]{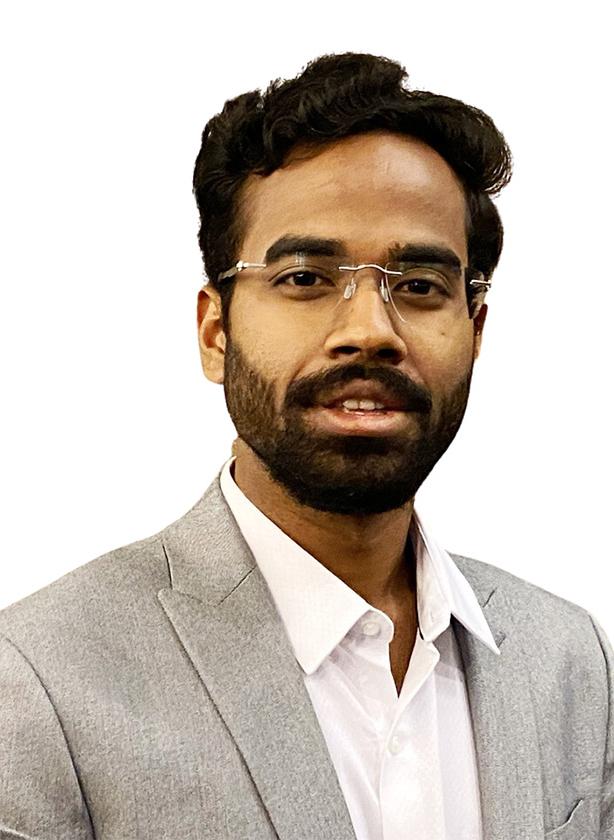}}]{Ayush Mohanty}
Ayush Mohanty is a Ph.D. student in Machine Learning at the School of Industrial and Systems Engineering, Georgia Institute of Technology, advised by Dr.~Nagi Gebraeel. He holds a B.S. in Manufacturing Science and Engineering and an M.S. in Industrial Engineering and Management from the Indian Institute of Technology (IIT) Kharagpur, India. His research focuses on decentralized learning of operational interdependencies in complex engineering systems, emphasizing federated learning frameworks for causal reasoning. As part of NASA’s HOME-STRI initiative, he has led demonstrations of self-aware deep-space habitats enabling real-time fault diagnostics and autonomous decision-making. His broader interests include prognostics for robotic manipulators and failure prediction in multi-fault environments.
\end{IEEEbiography}
\begin{IEEEbiography}[{\includegraphics[width=1in,height=1.15in,clip]{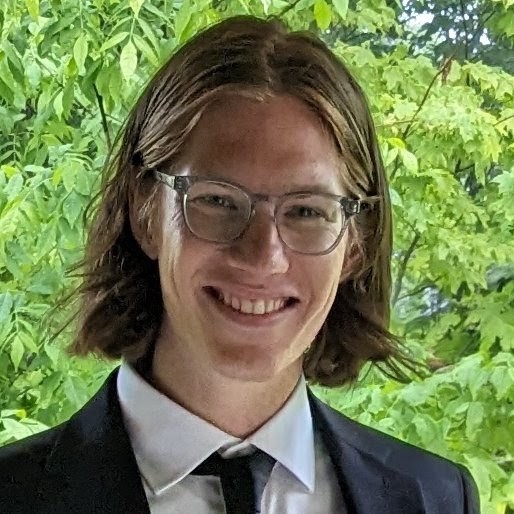}}]{Jason Dekarske}
Jason Dekarske is a Space Operations Engineer at Space Exploration Technologies Corp. He received his B.S. in Biomedical Engineering from the University of Wisconsin–Madison in 2018 and his Ph.D. in Mechanical and Aerospace Engineering from the University of California, Davis, in 2024. His research interests span robotics, controls, and space systems engineering, with applications in autonomous operations and fault-tolerant spacecraft design.
\end{IEEEbiography}
\begin{IEEEbiography}[{\includegraphics[width=1in,height=1.15in,clip]{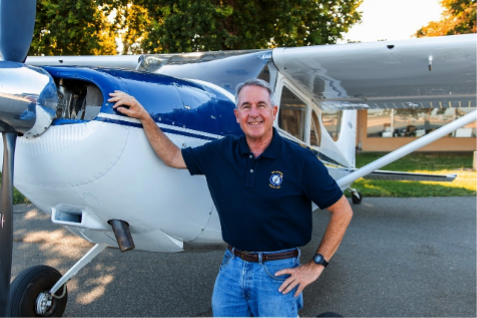}}]{Stephen K. Robinson}
Stephen K. Robinson is a Professor of Mechanical and Aerospace Engineering at the University of California, Davis, and the Director of the UC Davis Center for Space Exploration Research. Prior to joining academia, he spent 37 years at NASA as a technician, engineer, research scientist, pilot, and astronaut. During his 17-year astronaut career, he flew on four space shuttle missions, performed three spacewalks, and visited the International Space Station twice. His research interests include spacecraft design for human safety, human–robot teaming, AI-based spacecraft autonomy, and operational safety in space systems. Dr.~Robinson holds engineering degrees from UC Davis (B.S.), Stanford University (M.S. and Ph.D.), and has conducted research at Princeton and MIT.
\end{IEEEbiography}
\begin{IEEEbiography}[{\includegraphics[width=1in,height=1.15in,clip]{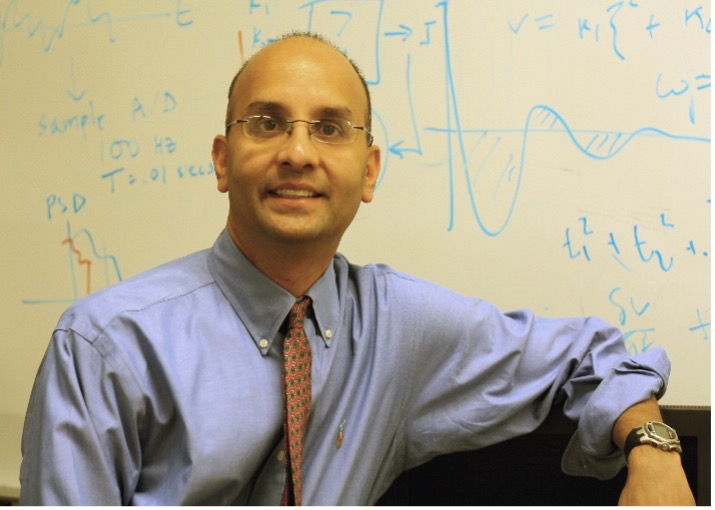}}]{Sanjay Joshi}
Sanjay Joshi is a Professor of Mechanical and Aerospace Engineering at the University of California, Davis, where he directs the Robotics, Autonomous Systems, and Controls Laboratory. His research interests include robotics, controls, space systems, and neuroengineering. He is a Senior Member of the IEEE.
\end{IEEEbiography}
\begin{IEEEbiography}[{\includegraphics[width=1in,height=1.15in,clip]{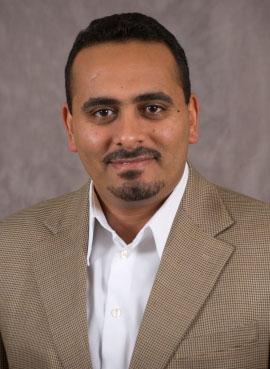}}]{Nagi Gebraeel}
Nagi Gebraeel (Member, IEEE) received the Ph.D. degree in Industrial Engineering from Purdue University, West Lafayette, IN, USA, in 2003. He is a Professor in the H.~Milton Stewart School of Industrial and Systems Engineering at the Georgia Institute of Technology, Atlanta, GA, USA. He serves as the Associate Director of Georgia Tech’s Strategic Energy Institute and the Director of the Analytics and Prognostics Systems Laboratory at the Georgia Tech Manufacturing Institute. His research focuses on predictive analytics and decision optimization models for large-scale industrial systems, with applications in energy and manufacturing. He is a member of INFORMS and a former President of the IISE Quality and Reliability Engineering Division.
\end{IEEEbiography}
\end{document}